\documentclass[letterpaper,10 pt,conference]{ieeeconf}  

\IEEEoverridecommandlockouts
\pdfoutput=1

\usepackage[noadjust]{cite}

\usepackage{xparse} 
\usepackage{amsmath} 
\usepackage{amssymb}
\usepackage{amsfonts} 
\usepackage{bbm}
\usepackage{color}
\usepackage{mathtools}
\usepackage{multirow}
\usepackage{graphicx}
\usepackage{subcaption}
\usepackage{calc}
\usepackage{balance}

\usepackage{booktabs}

\usepackage{url}

\usepackage{algpseudocode}
\usepackage{algorithm}

\makeatletter
\let\NAT@parse\undefined
\makeatother
\usepackage[colorlinks]{hyperref}
\usepackage[capitalize]{cleveref}

\usepackage{xparse}

\NewDocumentCommand\bbm{}{ \begin{bmatrix} }
\NewDocumentCommand\ebm{}{ \end{bmatrix} }

\NewDocumentCommand\Matrix{m}{ \boldsymbol{\mathbf{#1}} }

\NewDocumentCommand\LieGroupSE{m}{ \mathrm{SE}(#1) }

\NewDocumentCommand\CoordinateFrame{m}{ \underrightarrow{\Matrix{\mathcal{F}}}_{#1} }

\NewDocumentCommand\Transform{}{ \Matrix{T} }

\title{\LARGE\bf Self-Calibration of Mobile Manipulator Kinematic and Sensor\\ Extrinsic Parameters Through Contact-Based Interaction}
\author{Oliver Limoyo$^\dag$, Trevor Ablett$^\dag$, Filip Mari\'{c}$^\dag$$^\ddag$, Luke Volpatti$^\dag$, and Jonathan Kelly$^\dag$
	\thanks{$^\dag$All authors are with the Space \& Terrestrial Autonomous Robotic Systems (STARS) Laboratory at the University of Toronto Institute for Aerospace Studies (UTIAS), Toronto, Canada {\tt <firstname>.<lastname>@robotics.utias.utoronto.ca}.} 
    \thanks{$^\ddag$ F. Mari\'{c} is jointly with the Laboratory for Autonomous Systems and Mobile Robotics (LAMOR) at the University of Zagreb, Croatia.}}
\begin{document}
\maketitle
\thispagestyle{empty}
\pagestyle{empty}

\begin{abstract}
We present a novel approach for mobile manipulator self-calibration using contact information. Our method, based on point cloud registration, is applied to estimate the extrinsic transform between a fixed vision sensor mounted on a mobile base and an end effector. Beyond sensor calibration, we demonstrate that the method can be extended to include manipulator kinematic model parameters, which involves a non-rigid registration process. Our procedure uses on-board sensing exclusively and does not rely on any external measurement devices, fiducial markers, or calibration rigs. Further, it is fully automatic in the general case. We experimentally validate the proposed method on a custom mobile manipulator platform, and demonstrate centimetre-level post-calibration accuracy in positioning of the end effector using visual guidance only. We also discuss the stability properties of the registration algorithm, in order to determine the conditions under which calibration is possible.
\end{abstract}

\section{Introduction}

Collaborative robots, or \emph{cobots}, are machines designed to work alongside humans in shared spaces. Many cobots are mobile, and incorporate lightweight, person-safe robotic arms for manipulation tasks. These mobile manipulators need to be correctly calibrated to ensure accurate and reliable operation. Depending on the platform, this may involve determining intrinsic (e.g., camera focal length) and extrinsic (i.e., relative pose) sensor parameters, as well as kinematic parameters of the manipulator arm (e.g., joint biases and link length offsets). Additionally, many calibration parameters will change over a robot's lifetime due to general wear and tear, for example. One way to compensate for these changes is to employ self-calibration techniques, in which the system calibrates independently using only its on-board hardware. Robust self-calibration has already been demonstrated for certain sensor combinations, including cameras, inertial measurement units, and lidars (e.g., \cite{Nicholas_Roy1998-aw,Kelly2011-my,Sheehan2011-od,lambert2016entropy}). 

Despite the success of self-calibration for \emph{sensing} applications, there has been relatively little work on combined sensor-actuator self-calibration for mobile platforms. While classical manipulator kinematic model calibration has a long history in industrial environments \cite{mooring1991fundamentals}, these methods typically require specialized external measurement devices and human intervention. Similarly, much of the work on sensor (e.g., camera) extrinsic calibration relative to an arm's end-effector (EE) uses external fiducial markers \cite{pradeep2014calibrating} or specialized hardware attachments \cite{pastor2013learning} or equipment \cite{hu2013automatic}. The need for supplementary equipment, often not readily available in collaborative environments, makes these approaches less attractive than self-calibration.
\begin{figure}
\vspace{4mm}
\centering
\includegraphics[width=0.99\columnwidth]{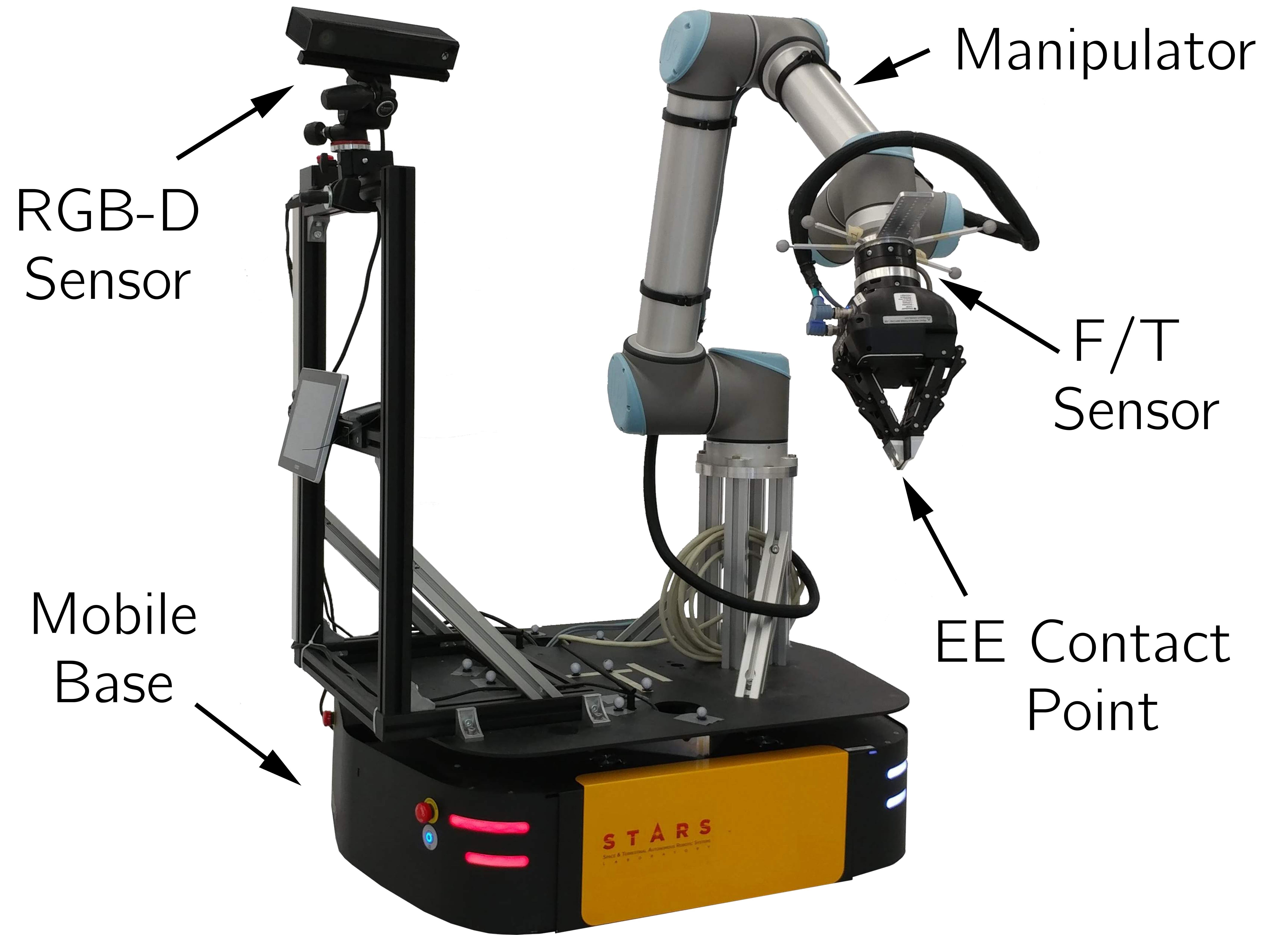}
\caption{Robot used for experimental evaluation. The platform incorporates a UR-10 manipulator mounted on a mobile base,\hspace{-0.2em} an RGB-D sensor, and a F/T sensor attached at the wrist of the manipulator.}
\label{fig:thing}
\vspace{-\baselineskip}
\end{figure}

\begin{figure}
   	\centering
   	\includegraphics[width=\columnwidth]{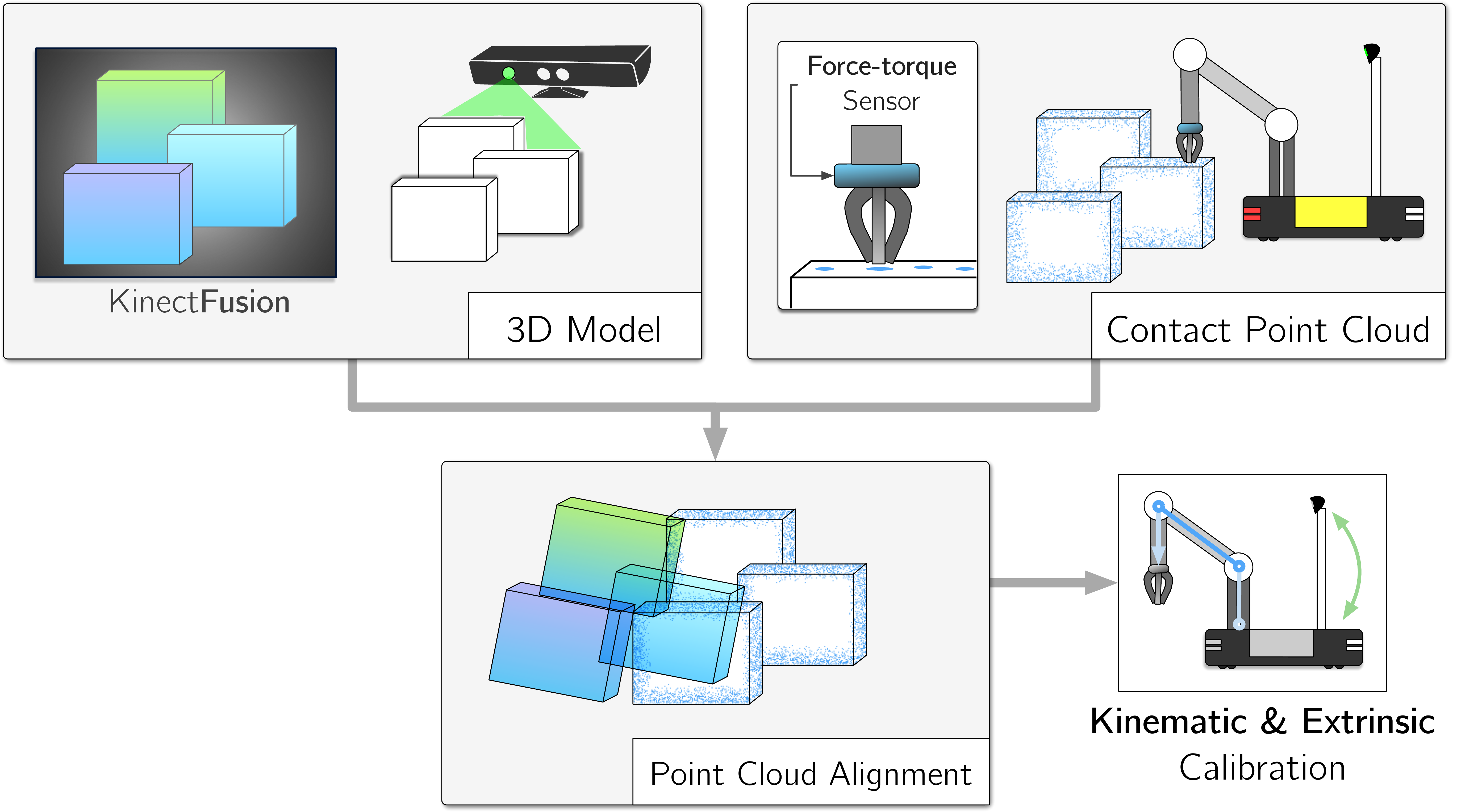}
   	\caption{Our method aligns two point clouds, derived from a depth camera and a contact sensor, to calibrate the camera extrinsics (and possibly manipulator kinematics) using ICP-based point cloud registration.}
   	\label{fig:system_overview}
   	\vspace{-\baselineskip}
\end{figure}

In this work, we propose a novel contact-based self-calibration solution for determining the extrinsic transform between a manipulator's EE and a fixed sensor (not attached to the manipulator itself) that is able to provide depth information. Our approach, summarized in \cref{fig:system_overview} and as shown in the accompanying video, does not make use of any markers or extra equipment; rather, we leverage the structure of the local environment (surfaces) for calibration. We generate a fused point cloud map from the depth sensor data by moving the mobile base to multiple vantage points. This point cloud map is then aligned with a \emph{contact point cloud map} (or contact map) produced by moving the EE over the same surfaces observed by the depth sensor. The contact map is generated by maintaining a fixed force profile (using a force-torque, or F/T, sensor near the EE) while following the surface contours. The extrinsic parameters are then recovered by aligning or registering the two point clouds using the iterative closest point (ICP) algorithm. We also demonstrate that kinematic model parameters can be introduced into the procedure.

This approach has a myriad of advantages: for example, calibration can be performed in situations where the robot's arm(s) may partially or completely occlude the depth sensor's field of view, or in cases where the EE itself is not visible at all. Because no external hardware is needed (other than a rigid 3D surface), the procedure can be performed online, as necessary, without the need to remove the robot from service. We note that the hardware  configuration described herein is very common; many mobile manipulators incorporate some type of contact or force sensor, while RGB-D cameras are becoming ubiquitous due to their cost and performance. Specifically, we make the following contributions:
\begin{enumerate}
	\item we develop an algorithm, based on point cloud registration, for contact-based calibration of the extrinsic parameters of a fixed depth camera not attached to the manipulator arm,
	\item we show that this algorithm can be extended to incorporate manipulator kinematic model parameters, through the use of non-rigid registration,
	\item we examine, via experiments, the observability and stability properties of ICP for the sensor-manipulator system, to determine the conditions under which extrinsic calibration is possible, and 
	\item we present a series of proof-of-concept results that demonstrate the accuracy of the proposed approach.
\end{enumerate}

\section{Related Work}

Manipulator-camera extrinsic calibration has been well studied in the context of ``eye-in-hand" systems (with the camera attached to the EE), and for fixed external cameras. For both configurations, the majority of existing calibration techniques make use of fiducial markers to facilitate EE localization \cite{kahn2014hand, birbach2015rapid}. In the ``eye-in-hand" case, the manipulator's EE motion is coupled with the camera's motion, allowing for the use of structure from motion techniques \cite{heller2011structure, schmidt2005calibration} to recover the transform. The work in \cite{birbach2015rapid} is related to our own, in that the transform of a fixed camera is determined using fiducial markers on the EE. However, for this approach to work, the EE must remain within the camera's field of view at all times.

Manipulator kinematic model calibration generally uses high-accuracy external measurement devices to track the EE \cite{nubiola2014absolute}, \cite{gaudreault2016local}, yielding sub-millimetre accuracy in some cases. We do not expect our accuracy numbers to be competitive with these techniques---however, we argue a) that our target applications, such as object grasping, typically do not require such accuracy, and b) that we have the advantage of being able to automatically calibrate without additional equipment in a large variety of environments.

\begin{figure}
	\vspace{4mm}
	\centering
	\includegraphics[width=0.85\columnwidth]{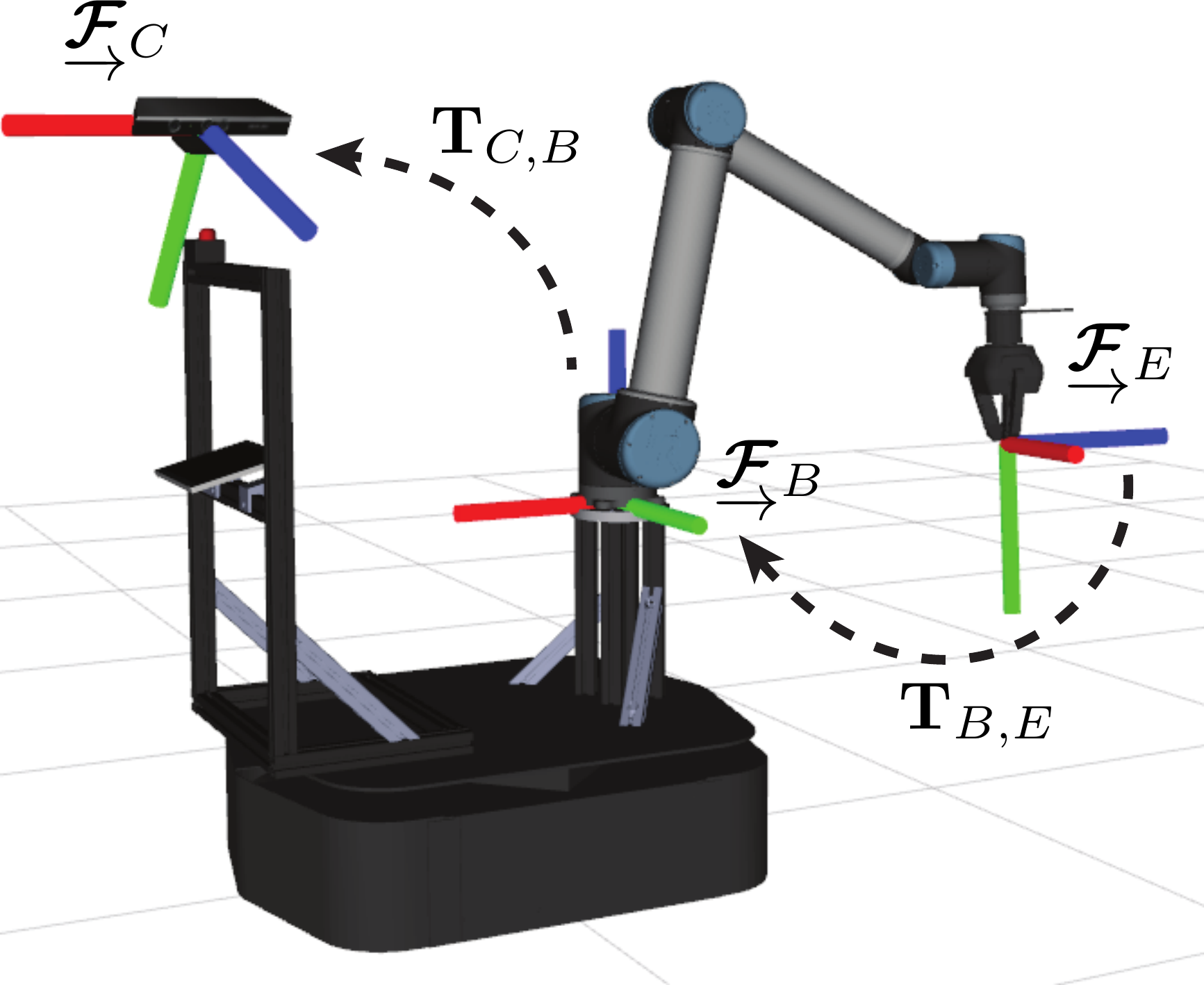}
	\caption{Reference frames for the calibration task. $\CoordinateFrame{C}$ is the optical frame of the depth sensor (camera), $\CoordinateFrame{B}$ is the manipulator's base frame, and $\CoordinateFrame{E}$ is the frame located at the tip of the EE, where contact with a surface is made. The transform $\Transform_{B,E}$ is given by the forward kinematics of the arm, while $\Transform_{C,B}$ is the extrinsic transform that we seek to determine.}
	\label{fig:frames}
	\vspace{-0.74\baselineskip}
\end{figure}

The use of contact (or touch) in the context of perception, motion planning, and manipulation has garnered significant attention recently. In \cite{Yu2015-zl}, the authors exploit the local but detailed nature of contact measurements to recover the shape and pose of a movable planar object, drawing inspiration from simultaneous localization and mapping (SLAM) algorithms. Similarly, contact has been combined with vision as part of a multi-modal strategy for tracking objects \cite{izatt2017tracking, Koval2013-nh}. In these systems, contact plays a complementary role to vision by providing information at the time of grasping, when the robot's manipulator most likely occludes the object. 

The observability, convergence, and accuracy of point cloud registration \cite{besl1992method, chen1991object} has been explored by both the robotics and computer graphics communities. In \cite{Censi2007-km} and \cite{Bonnabel2014-mo}, derivations of the observability properties of the point-to-point and point-to-plane ICP cost functions are presented and shown to be related to the eigenvalues and eigenvectors of the corresponding approximate Hessian matrices. Similarly, in \cite{Gelfand2003-ar} and \cite{Rusinkiewicz2001-jg}, a study of the observability of ICP when sampling from common surface classes is carried out, and the respective unconstrained or unobservable directions of surface motion (`sliding') are identified. The quality of registration is related to the choice of points used, and a sampling strategy is proposed that ensures stable registration. In \cite{Brown2007-tk}, non-rigid ICP is employed to more accurately align two 3D point clouds by considering camera calibration errors as inducing non-rigid cloud deformations. In our work, we consider kinematic model errors in the arm, as opposed to camera intrinsic mis-calibration, as a source of non-rigid deformations. 
 
\section{Extrinsic Calibration from Depth\\ and Contact}
\label{sec:theory}

As an initial formulation of the problem, we consider calibration of the extrinsic transform only, without kinematic model parameters. We solve for the transform $\Transform_{C,B} \in \LieGroupSE{3}$, between the manipulator's base frame $\CoordinateFrame{B}$ and the depth camera's optical frame $\CoordinateFrame{C}$, as shown in \cref{fig:frames}. The transform  parameters are
\begin{align}
	\boldsymbol{\Xi} &= \bbm x_{e} & y_{e} & z_{e} & \phi_{e} & \theta_{e} & \psi_{e} \ebm ^{T},
\end{align}

\noindent where $\boldsymbol{\Xi}$ includes three translation and three rotation parameters. We assume access to an intrinsically calibrated depth camera capable of generating a 3D point cloud map and also assume that contact in the EE's frame $\CoordinateFrame{E}$ can be detected and estimated as a 3D point measurement. Further, we assume that all surfaces are rigid and that there is negligible deformation during contact.

The contact sensor provides a point cloud map $\mathbf{A}$ in $\CoordinateFrame{B}$, while the depth camera provides a second point cloud map $\mathbf{B}$ in $\CoordinateFrame{C}$, 
\begin{align}
	\mathbf{A} &= \{\mathbf{a}_{1}, \mathbf{a}_{2}, \hdots, \mathbf{a}_{n} \},\hspace{2mm} 
    \mathbf{B} = \{ \mathbf{b}_{1}, \mathbf{b}_{2}, \hdots, \mathbf{b}_{m} \},
\end{align}
\noindent where $\mathbf{a}_{i}$ and $\mathbf{b}_{j}$ are the 3D coordinates of points in the two clouds, respectively. We denote the homogeneous form of $\mathbf{a}_{i}$ (i.e., a $4 \times 1$ vector) as $\boldsymbol{a}_{i}$.

An important aspect to consider is that the contact point is located at the origin of $\CoordinateFrame{E}$, which is a moving frame following the EE's trajectory. In order to generate a consistent contact map, the points $\mathbf{A}$ need to be represented in the \emph{fixed} frame $\CoordinateFrame{B}$. The transform $\Transform_{B,E}(\boldsymbol{\theta}_{i}, \boldsymbol{\Psi})$ between the manipulator's base frame and the EE is assumed to be known from the corresponding $K$ joint encoder readings, $\boldsymbol{\theta}_{i} = \bbm \theta_{1,i}, \hdots, \theta_{K,i} \ebm$, for each respective contact point $\mathbf{a}_{i}$. Given a kinematic model with known nominal parameters, $\boldsymbol{\Psi} = \{\boldsymbol{\psi}_{1}, \hdots, \boldsymbol{\psi}_{K}\}$, each contact point is represented in the base frame as
\begin{align} \label{eq:contact_point_def}
	\boldsymbol{a}_{i} = \Transform_{B,E}(\boldsymbol{\theta}_{i}, \boldsymbol{\Psi}) \begin{bmatrix} 0 & 0 & 0 & 1 \end{bmatrix}^T.
\end{align}
We use the Denavit-Hartenberg (DH) \cite{denavit1955kinematic} parametrization for forward kinematics,
\begin{align}
	\begin{split}
		\Transform_{B,E}(\boldsymbol{\theta}_{i}, \boldsymbol{\Psi}) =\; & \mathbf{D}_{0,1}(\theta_{1,i}, \boldsymbol{\psi}_{1})\hdots \mathbf{D}_{k-1,k}(\theta_{k,i}, \boldsymbol{\psi}_{k}) \hdots \\
		& \mathbf{D}_{K-1, K}(\theta_{K,i}, \boldsymbol{\psi}_{K}),
	\end{split}
\end{align}
\noindent where each $\mathbf{D}_{k-1,k}$ is the respective DH matrix, from manipulator joint frame $k-1$ to $k$ with parameters $\boldsymbol{\psi}_{k} = \bbm \alpha_{k} \!& r_{k} &\! d_{k}\ebm$, given as
\begin{align}
	\mathbf{D}_{k-1,k} &=
	\begin{bsmallmatrix}
	\cos{\theta_{k}} & -\sin{\theta_{k}}\cos{\alpha_{k}} & \sin{\theta_{k}}\sin{\alpha_{k}} & r_{k}\cos{\theta_{k}} \\
	\sin{\theta_{k}} & \cos{\theta_{k}} \cos{\alpha_{k}} & -\cos{\theta_{k}}\sin{\alpha_{k}} & r_{k}\sin{\theta_{k}} \\
	0 & \sin{\alpha_{k}} & \cos{\alpha_{k}} & d_{k}\\
	0 & 0 & 0 & 1\\
	\end{bsmallmatrix}.
\end{align}
The full transform between the EE and depth camera frames is then
\begin{align}
\Transform_{C,E}(\boldsymbol{\Xi}, \boldsymbol{\theta}_{i}, \boldsymbol{\Psi}) = \Transform_{C,B}(\boldsymbol{\Xi})\,\Transform_{B,E}(\boldsymbol{\theta}_{i}, \boldsymbol{\Psi}).
\end{align}

\subsection{Rigid ICP}

\begin{figure}
    \centering
    \setlength{\fboxsep}{0pt}%
    \setlength{\fboxrule}{1pt}%
    \begin{subfigure}{0.49\columnwidth}
        \fbox{\includegraphics[width=\textwidth - 2pt]{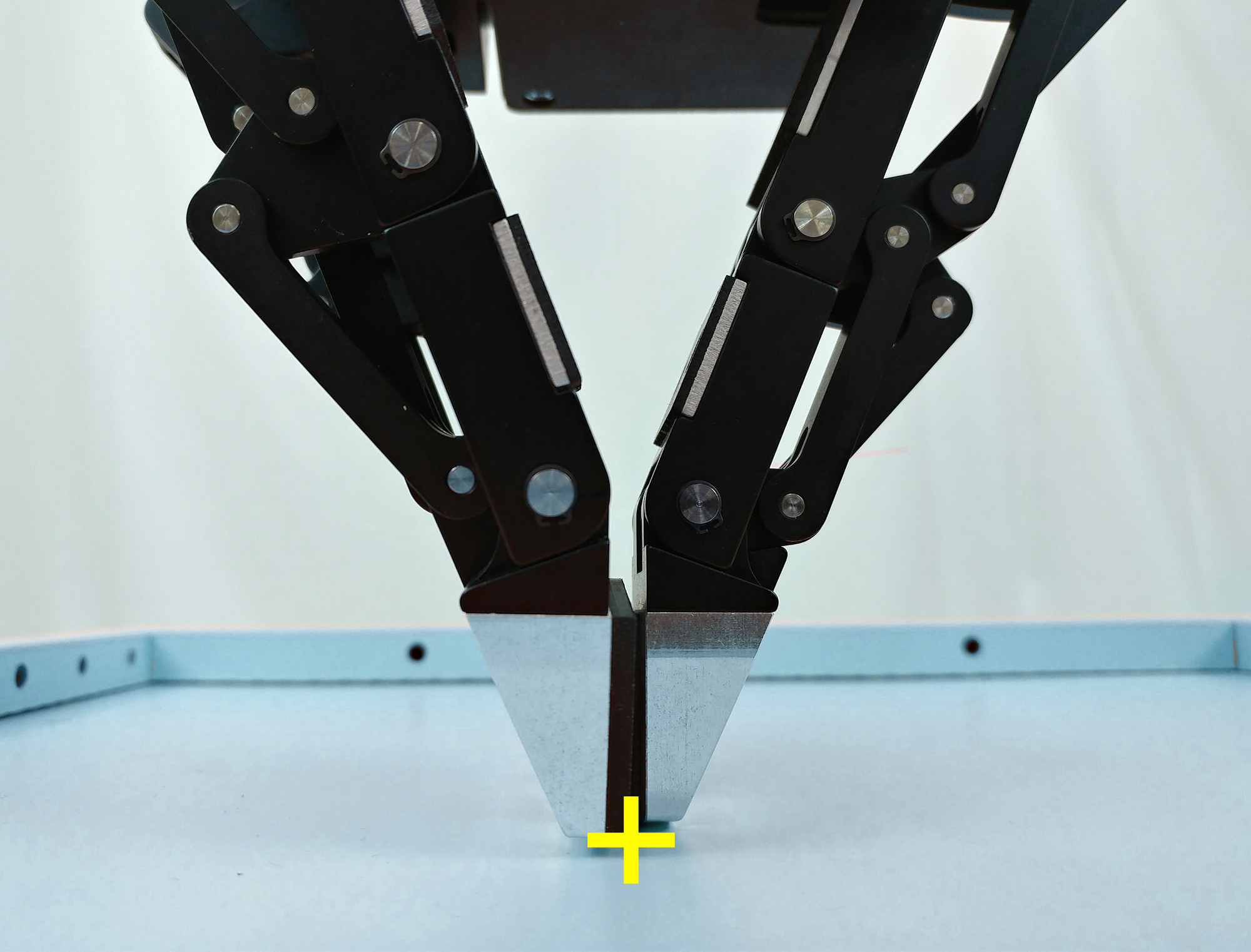}}
        \caption{Contact point front view.}
    \end{subfigure}
    \hfill
    \begin{subfigure}{0.49\columnwidth}
        \fbox{\includegraphics[width=\textwidth - 2pt]{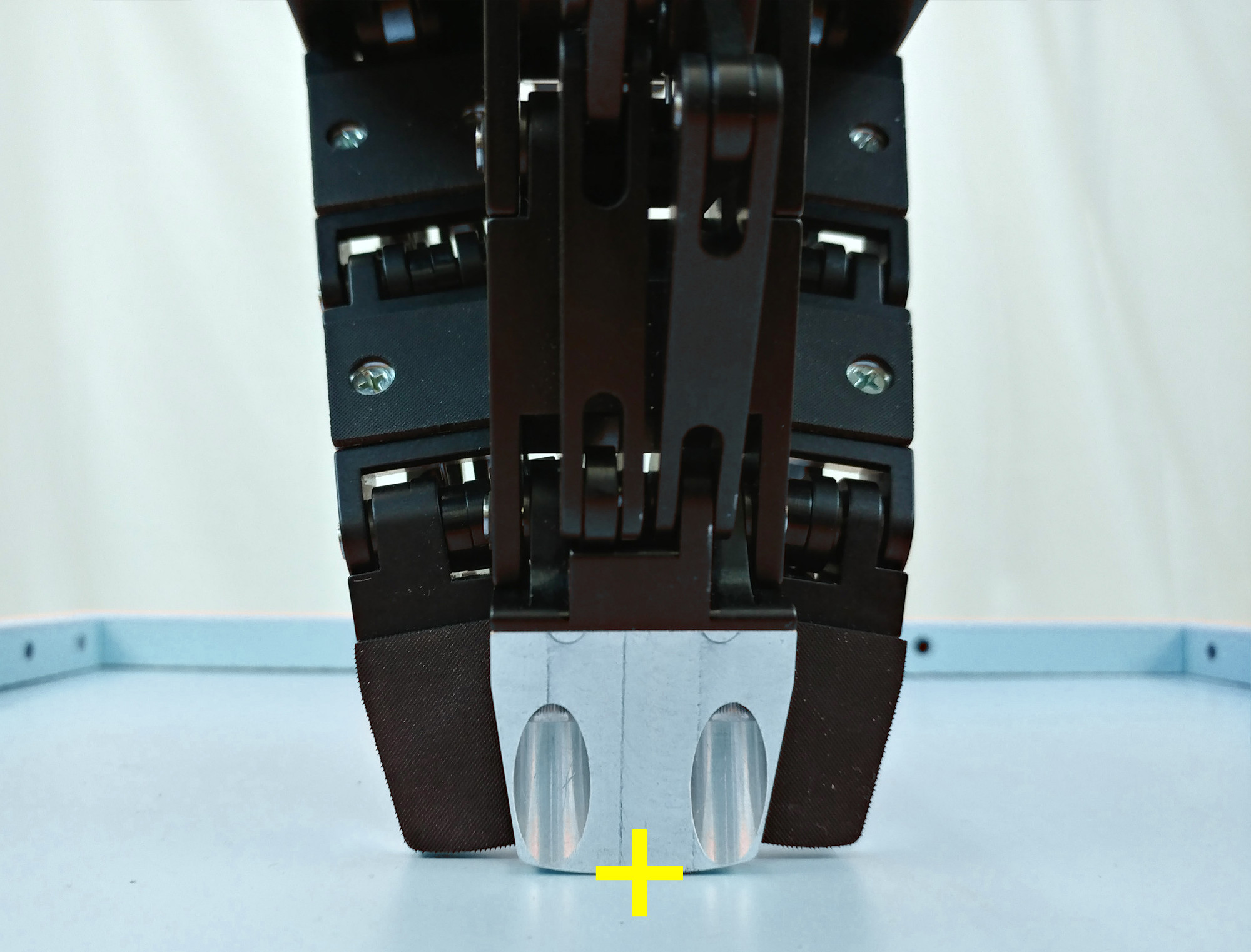}}
        \caption{Contact point side view.}
    \end{subfigure}
    \caption{A visualization of the location of the contact point at our EE's tip, overlaid as a yellow cross.}
    \label{fig:contact}
    \vspace{-\baselineskip}
\end{figure}

For extrinsic calibration, we use the point-to-plane variant of the ICP algorithm to register the two point clouds. We chose the point-to-plane error metric \cite{chen1991object} for our cost function $J_{pn}$ in order to best leverage the surface information contained in the dense depth sensor map. The error function to be minimized is then, explicitly,
\begin{align}
\label{eq:error}
	 J_{pn}(\boldsymbol{\Xi}) &= \sum_{i} w_{i} \lVert \mathbf{n}_{i}^{T}(\mathbf{P} \Transform_{C,B}(\boldsymbol{\Xi})\,\boldsymbol{a}_{i} - \mathbf{b}_{i}) \rVert^{2},
\end{align}
\noindent where $\Transform_{C,B}$ is the rigid transform that we solve for, $\boldsymbol{a}_{i}$ are the contact points (in homogeneous form), defined by \cref{eq:contact_point_def}, $\mathbf{b}_{i}$ are the associated or paired depth camera points with their respective surface normals, $\mathbf{n}_{i}$, and $w_{i}$ are weights used for outlier removal. The matrix $\mathbf{P}$ is
\begin{align}
	\mathbf{P} = \bbm \mathbf{I}_{3} & \mathbf{0}_{3 \times 1} \ebm,
\end{align}
where $\mathbf{I}_{3}$ is the $3 \times 3$ identity matrix. Note that we use an idealized point estimate of the EE's flat tip, and that the point-to-plane metric does not consider the uncertainty in tangential motion of the EE's contact point (see \cref{fig:contact}). 

\subsection{Shape, Contact, and Observability}
\label{subsec:observability}

The choice of contact points to sample and the shapes of the surfaces in the environment affect the accuracy and convergence of our calibration solution. As shown in \cite{Gelfand2003-ar} and \cite{Rusinkiewicz2001-jg}, different sampling strategies and available surface shapes may result in instabilities during point cloud registration.
In our application, the sampling strategies outlined in \cite{Gelfand2003-ar} and \cite{Rusinkiewicz2001-jg} can be directly applied by having the robot actively pick specific contact areas. However, sampling is limited to the surfaces in the local environment. For example, if the environment consists of a single planar surface only, sampling any set of contact points will lead to underconstrained registration or sliding, as there is no change in registration cost in any direction parallel to the plane. 
Importantly, \cite{Gelfand2003-ar} demonstrates that sampling from three orthogonal planar surfaces is sufficient to constrain a rigid transform. 

Given the point-to-plane error metric, we can obtain a principled measure of the stability of a solution, for a specific set of contact points $\mathbf{A}$ and depth points in $\mathbf{B}$, by examining the eigenvalues of the approximate Hessian \cite{Rusinkiewicz2001-jg} of the linearized cost function. Linearizing \cref{eq:error} gives
\begin{equation}
\label{eq:error_jacobian}
  \bar{J}_{pn}(\Delta\boldsymbol{\Xi}) = \sum_{i} w_{i} \lVert r_{i} - \mathbf{J}_{i}\Delta\boldsymbol{\Xi} \rVert^{2},
\end{equation}
where $\Delta\boldsymbol{\Xi}$ is an incremental (small angle) update to the current transform parameters, $\boldsymbol{\Xi}$. The residuals $r_{i}$ and Jacobian matrices $\mathbf{J}_{i}$ are, respectively,
\begin{equation}
  r_{i} = \mathbf{n}_{i}^{T}(\mathbf{P} \Transform_{C,B}(\boldsymbol{\Xi})\,\boldsymbol{a}_{i}- \mathbf{b}_{i}),
\end{equation}
and 
\begin{equation}
  \mathbf{J}_{i} = \bbm -\mathbf{n}_{i}^{T} & -(\mathbf{a}_{i}^{\times} \mathbf{n}_{i})^{T} \ebm,
\end{equation}
where $\mathbf{a}_{i}^{\times}$ is the skew-symmetric matrix form of $\mathbf{a}_{i}$. In the vicinity of the true minimum, $\boldsymbol{\Xi} = \hat{\boldsymbol{\Xi}}$, following the approach in \cite{Bonnabel2014-mo} and solving \cref{eq:error_jacobian} for the incremental  update yields the following quadratic form,
\begin{equation}
\label{eq:cost_param_relationship}
  \Delta J_{pn}(\Delta\boldsymbol{\Xi}) = \Delta\boldsymbol{\Xi}^{T} \mathbf{Q} \Delta\boldsymbol{\Xi},
\end{equation}
where $\Delta J_{pn}=J_{pn}(\boldsymbol{\Xi})-J_{pn}(\hat{\boldsymbol{\Xi}})$, $\Delta\boldsymbol{\Xi}=\boldsymbol{\Xi}-\hat{\boldsymbol{\Xi}}$ and $\mathbf{Q}=\sum_{i} \mathbf{J}_{i}^{T}\mathbf{J}_{i}$. \cref{eq:cost_param_relationship} measures how the cost changes as $\boldsymbol{\Xi}$ moves away from the minimum $\hat{\boldsymbol{\Xi}}$. If a change in $\Delta\boldsymbol{\Xi}$ results in little (no) change in $\Delta J_{pn}$, then the solution is underconstrained (resp.\ unconstrained) in that direction. Further, a small eigenvalue of the approximate Hessian $\mathbf{Q}$ identifies an unobservable motions in the direction of the associated eigenvector. Thus, we choose our measure of stability or observability, as in \cite{Gelfand2003-ar}, to be based on the condition number $c$ of the matrix $\mathbf{Q}$,
\begin{align} \label{eq:stability_metric}
c &= \frac{\lambda_{1}}{\lambda_{6}},
\end{align}
where $\lambda_{1} > \lambda_{2} > \hdots > \lambda_{6}$ are the eigenvalues of $\mathbf{Q}$.

\section{Manipulator Kinematic Model Calibration Through Non-Rigid ICP}

It is possible to include bias parameters (i.e., joint angle biases, $\boldsymbol{\delta\theta}$, and other geometric biases, $\boldsymbol{\delta\Psi}$), within $\Transform_{B,E}$, in the ICP cost function. Instead of solving for the set of all possible DH biases, we consider joint angle biases ($\boldsymbol{\delta\theta}$) only in our initial derivation. 
Our implicit assumption is that joint angle biases will induce larger point cloud deformations due to moment arm effects. The resulting transform is no longer rigid; the homogeneous coordinates of contact points in the manipulator base frame are now defined as
\begin{equation}
\tilde{\boldsymbol{a}}_{i} = \Transform_{B,E}(\boldsymbol{\theta}_{i} + \boldsymbol{\delta\theta}, \boldsymbol{\Psi}) \begin{bmatrix} 0 & 0 & 0 & 1 \end{bmatrix}^T.
\end{equation}
The updated ICP cost function is given by 
\begin{equation}
	 \tilde{J}_{pn}(\boldsymbol{\Xi}, \boldsymbol{\delta\theta}) = \sum_{i} w_{i} \lVert \mathbf{n}_{i}^{T}(\mathbf{P}\Transform_{C,B}(\boldsymbol{\Xi})\,\tilde{\boldsymbol{a}}_{i} - \mathbf{b}_{i}) \rVert^{2},
\end{equation}
where $\tilde{\boldsymbol{a}}_{i}$ is the homogeneous form of $\tilde{\mathbf{a}}_{i}$. 
Therefore, $\tilde{J}_{pn}$ is a function of $6 + K$ parameters: six parameters that define the extrinsic transform, $\boldsymbol{\Xi}$, and an additional $K$ that form the set $\boldsymbol{\delta\theta}$ of joint angle biases for a $K$-DOF (rotary joint) manipulator. To determine $\boldsymbol{\Xi}$ and $\boldsymbol{\delta\theta}$, we use a standard nonlinear least squares solver (i.e., Levenberg-Marquardt).

While our method generalizes to complex manipulator configurations, redundant DOFs may lead to groups of trajectories that produce essentially the same contact map. Solving this more difficult problem could require more complex surface shapes, higher-fidelity contact sensing, and a wider range of motions. We leave further analysis as future work.

\begin{figure}
    \centering   
    \begin{subfigure}[t]{0.49\columnwidth}
        \includegraphics[width=\textwidth]{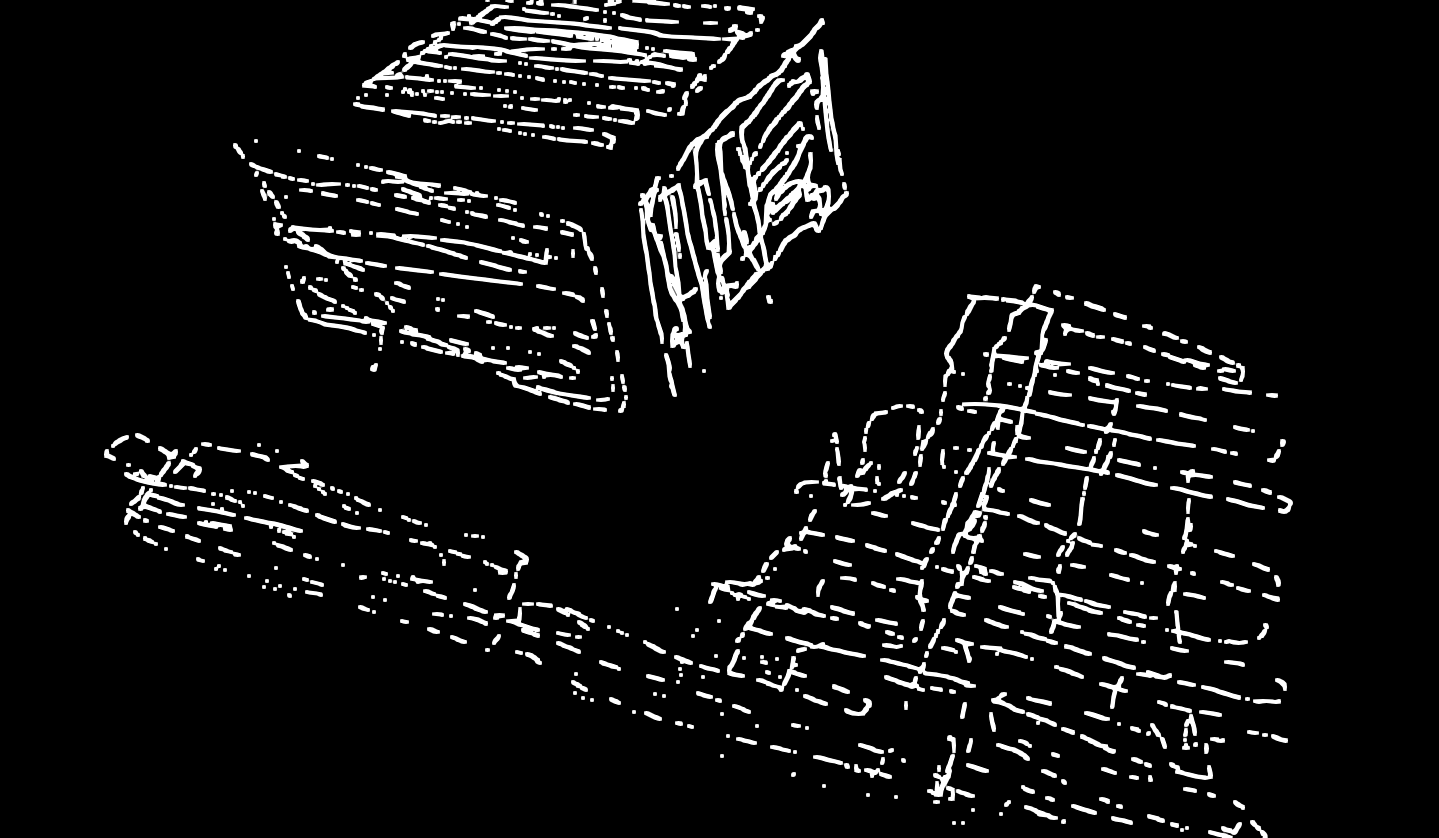}
    \end{subfigure}
    \hfill
    \begin{subfigure}[t]{0.49\columnwidth}
        \includegraphics[width=\textwidth]{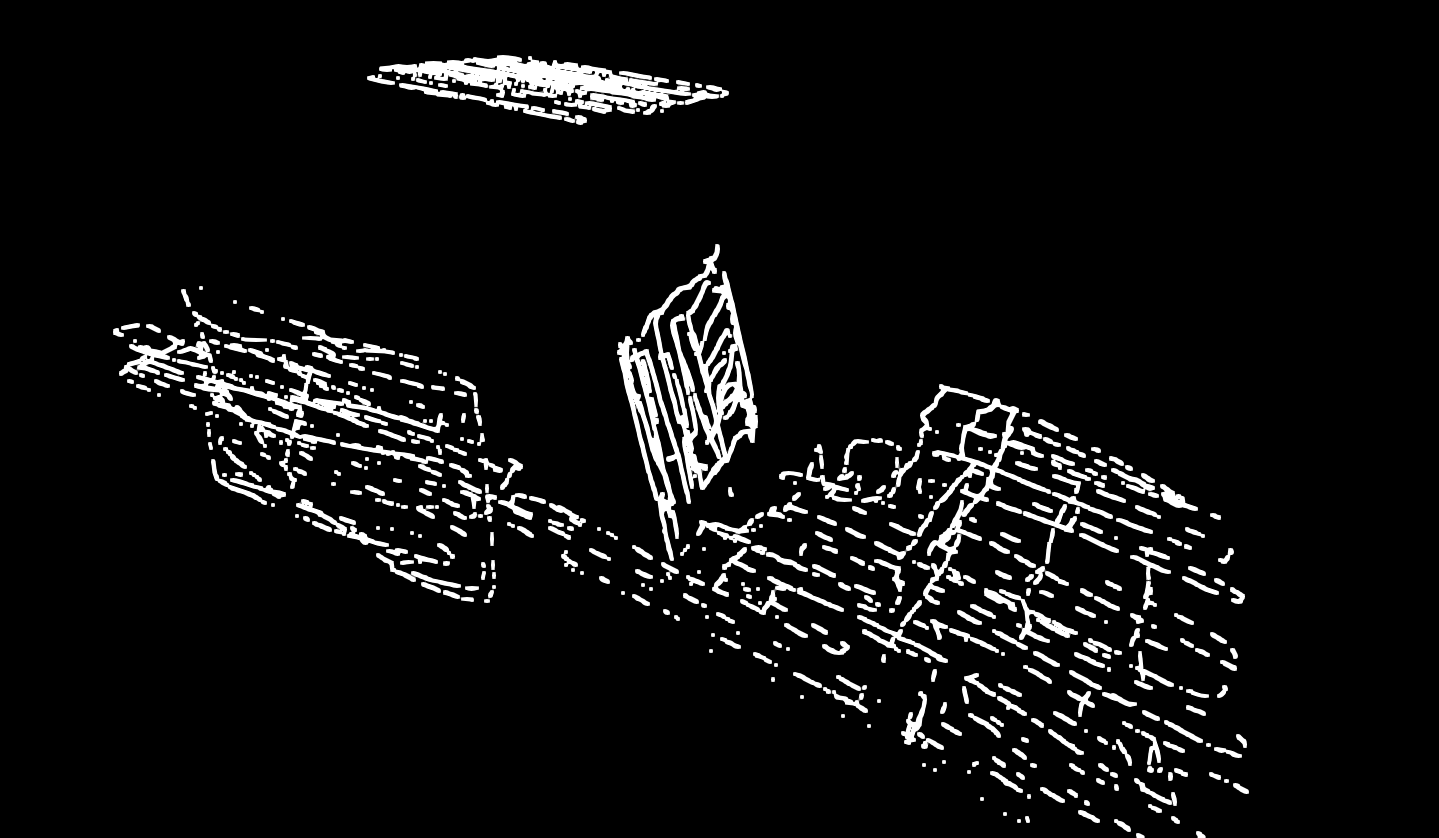}
    \end{subfigure}
    \par\vspace*{1.5mm} 
    \begin{subfigure}[t]{0.49\columnwidth}
        \includegraphics[width=\textwidth]{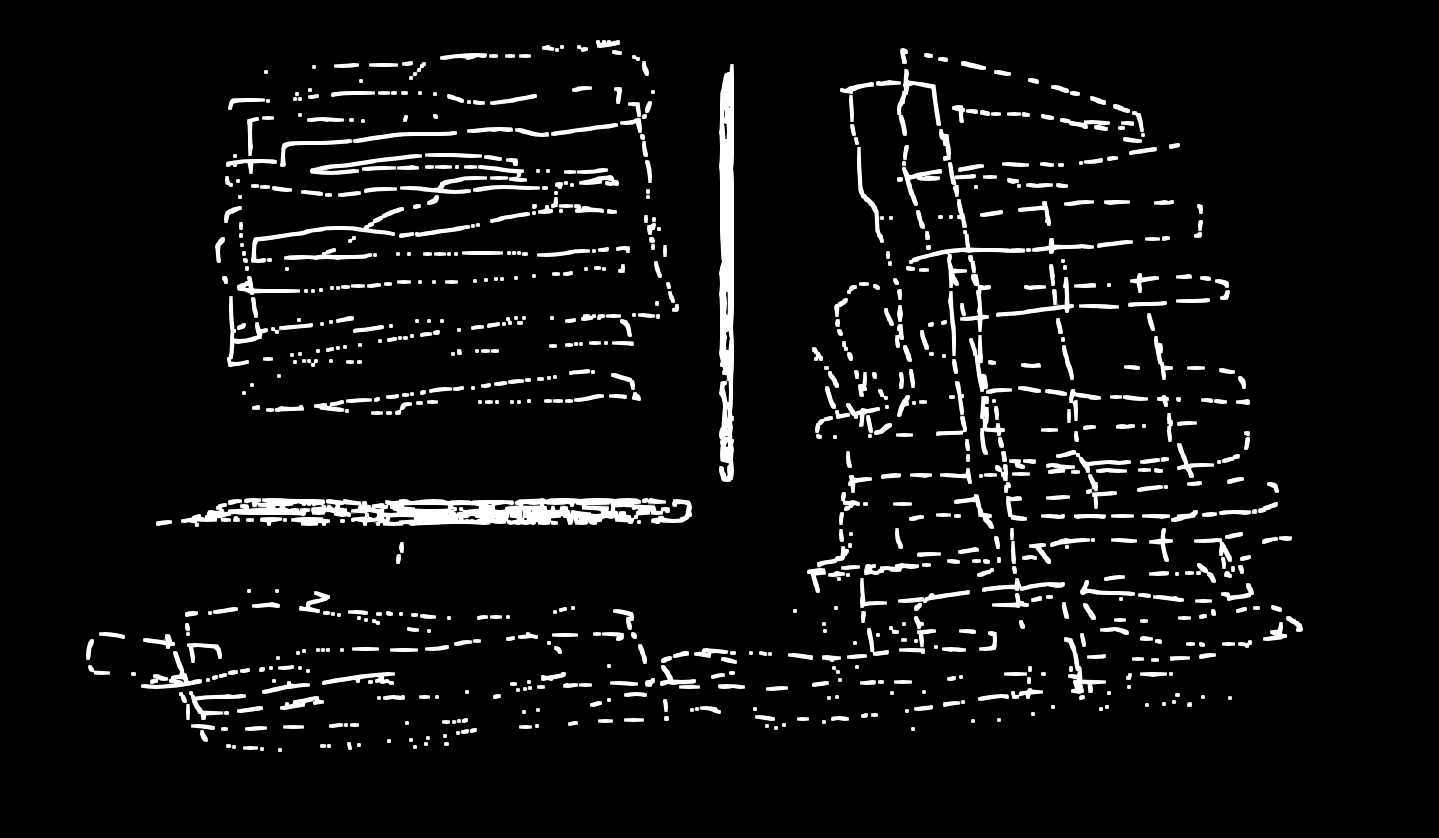}
    \end{subfigure}
    \hfill
    \begin{subfigure}[t]{0.49\columnwidth}
        \includegraphics[width=\textwidth]{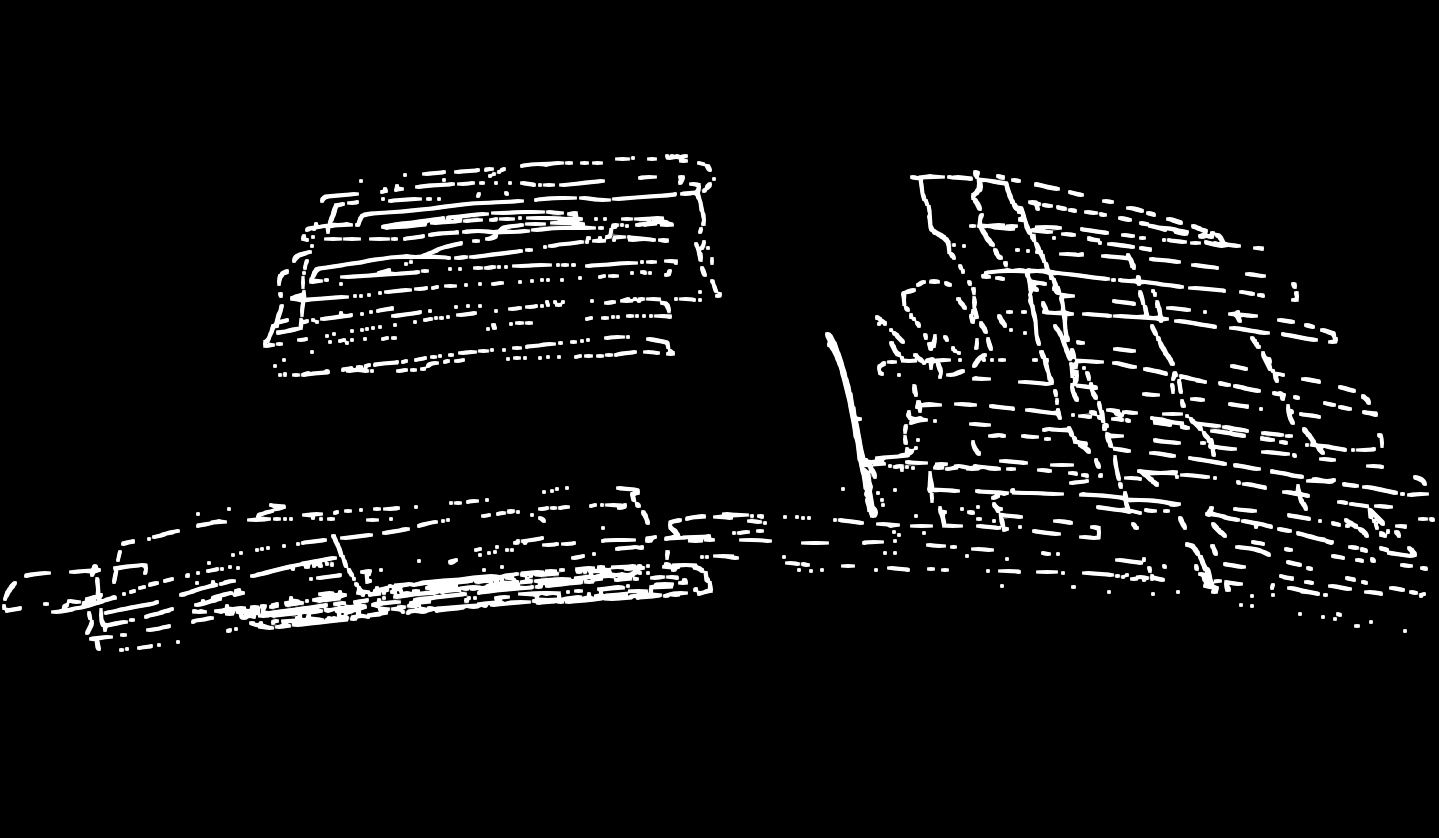}
    \end{subfigure}
    \par\vspace*{1.5mm} 
    \begin{subfigure}[t]{0.49\columnwidth}
        \includegraphics[width=\textwidth]{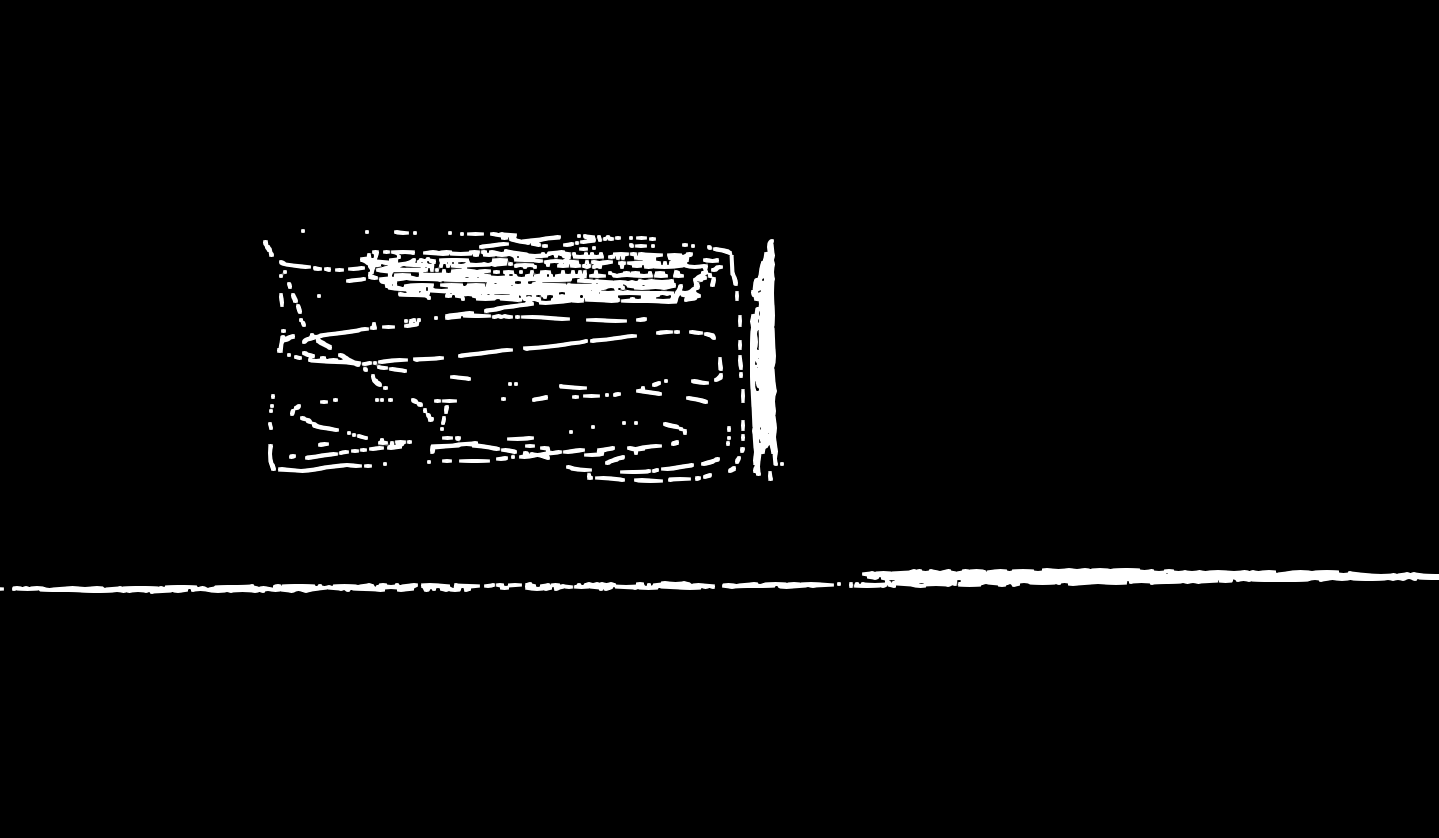}
    \end{subfigure}
    \hfill
    \begin{subfigure}[t]{0.49\columnwidth}
        \includegraphics[width=\textwidth]{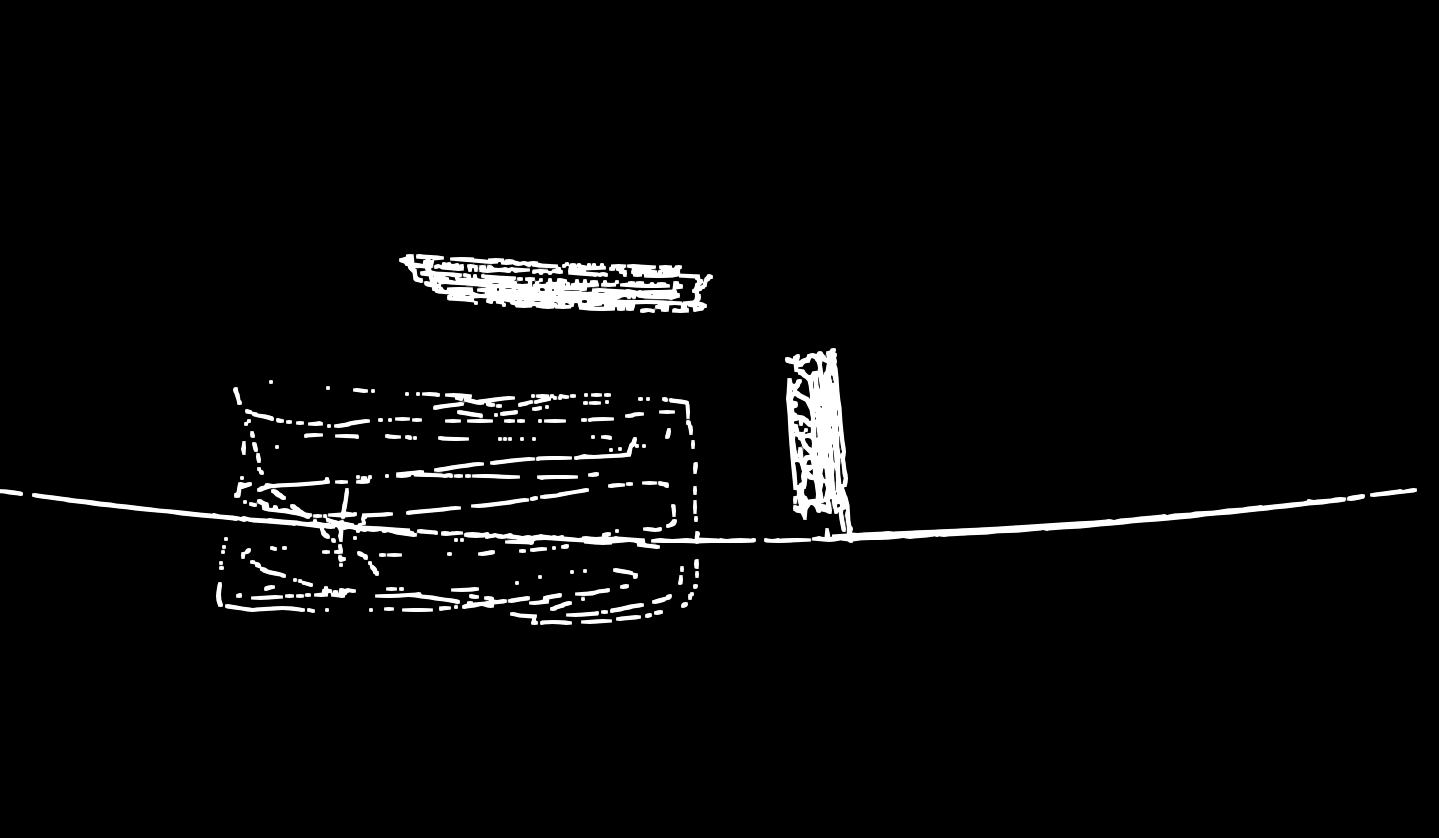}
    \end{subfigure}
    \caption{An example of the deformation of a contact map when a (simulated) bias of 29 degrees is added to Joint 4, the first wrist joint of our manipulator. Left column: Isometric, top, and front views of the true contact map. Right column: corresponding views with map deformation due to the joint bias.}
    \label{fig:biased_po	int_cloud}
    \vspace{-\baselineskip}
\end{figure}

Because the transform is no longer rigid, each point in the contact map will move individually as the parameters in $\boldsymbol{\delta\theta}$ are modified. The effects of a single (large) joint bias error on one contact map are shown in \cref{fig:biased_po	int_cloud}. Although the bias value (0.5 radians or 29 degrees) is unrealistic, we use this quantity to clearly demonstrate the effects of kinematic model errors on contact map deformation.

\section{Experiments and Results}
\label{sec:experiments}

\begin{figure*}
	\centering
	\setlength{\fboxsep}{0pt}%
    \setlength{\fboxrule}{1pt}%
	\begin{subfigure}{0.326\textwidth}
		\fbox{\includegraphics[width=\textwidth - 2pt]{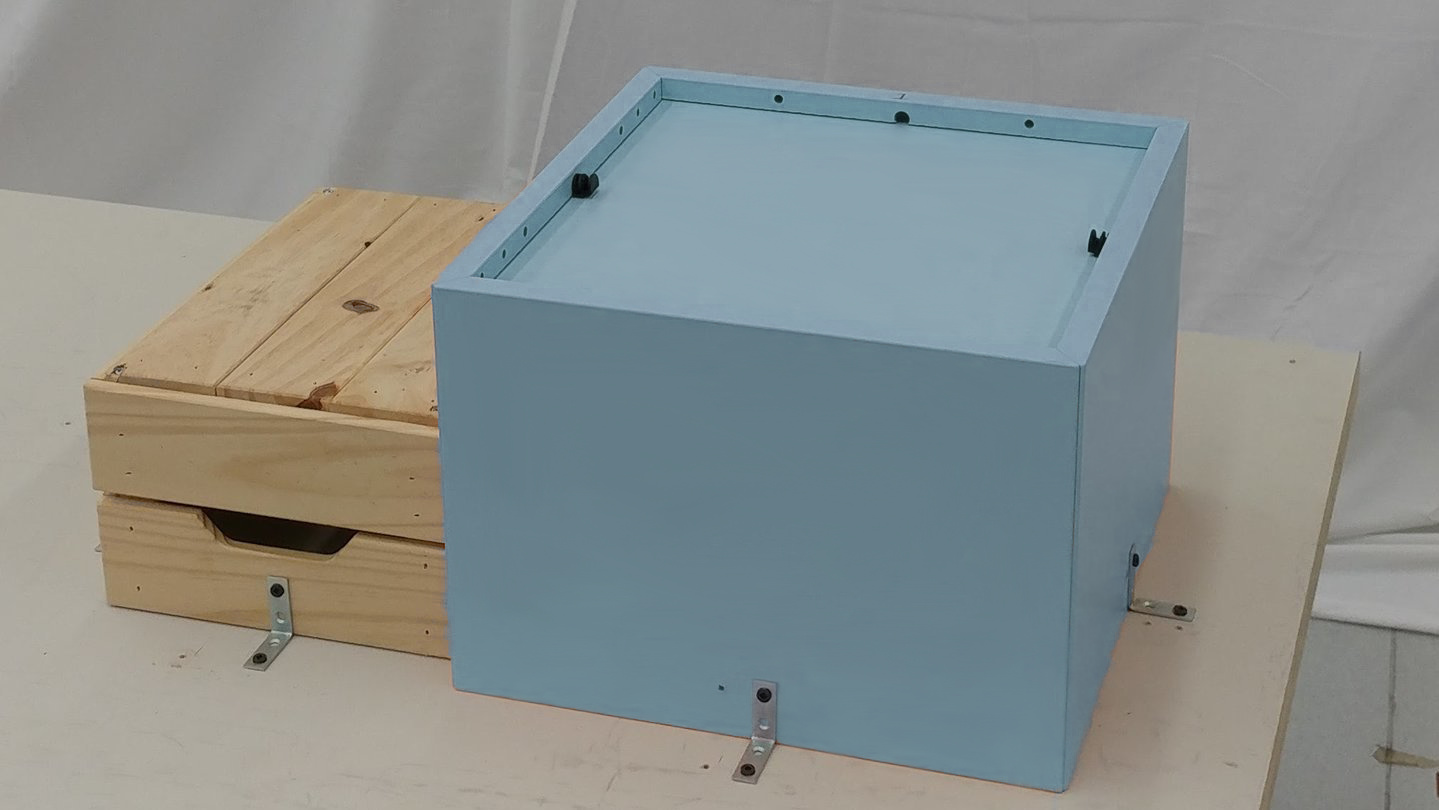}}
		\caption{Calibration surfaces (boxes).}
	\end{subfigure}
	\hfill
	\begin{subfigure}{0.326\textwidth}
		\fbox{\includegraphics[width=\textwidth - 2pt]{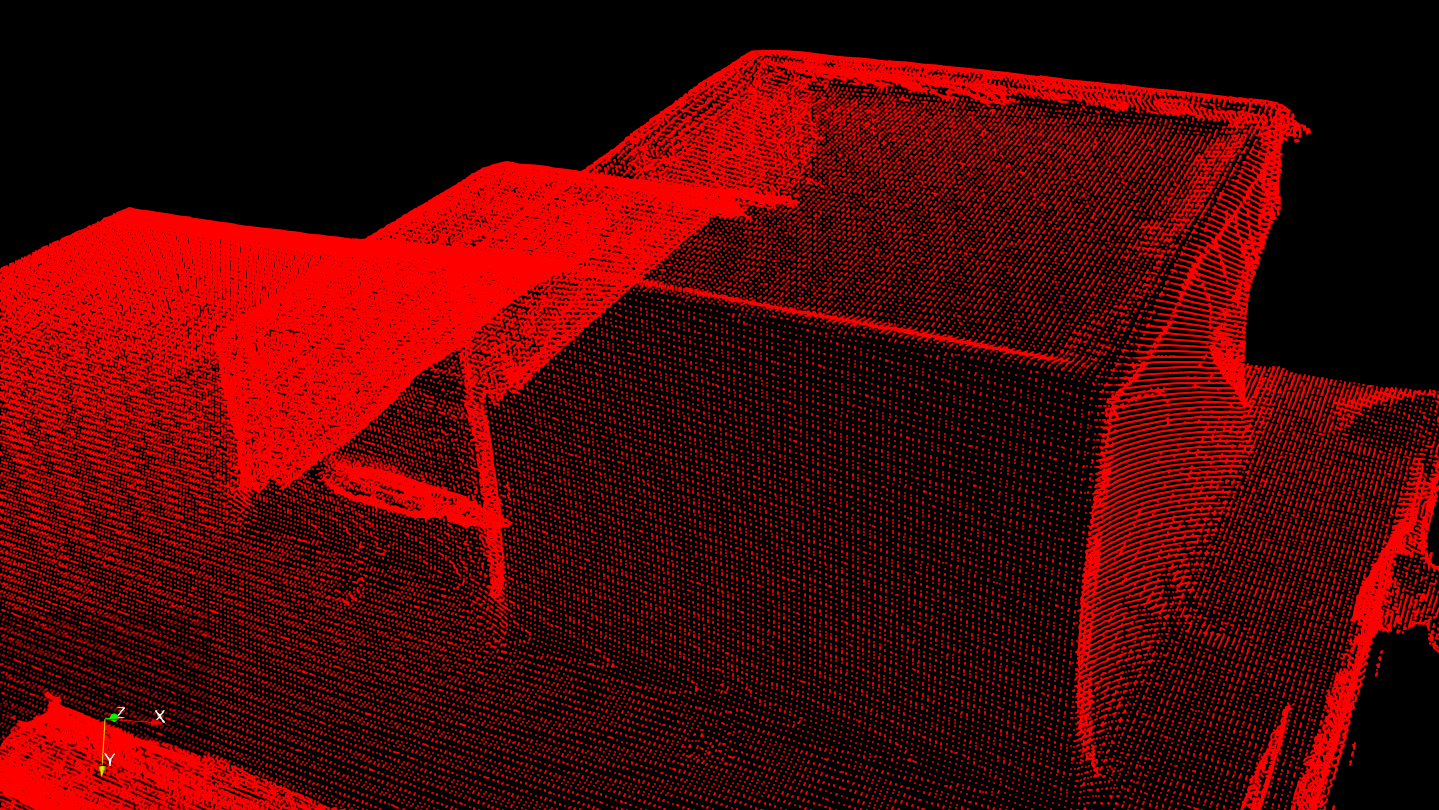}}
		\caption{Fused depth point cloud.}
		\label{fig:kinect_cloud}
	\end{subfigure}
	\hfill
	\begin{subfigure}{0.326\textwidth}
		\fbox{\includegraphics[width=\textwidth - 2pt]{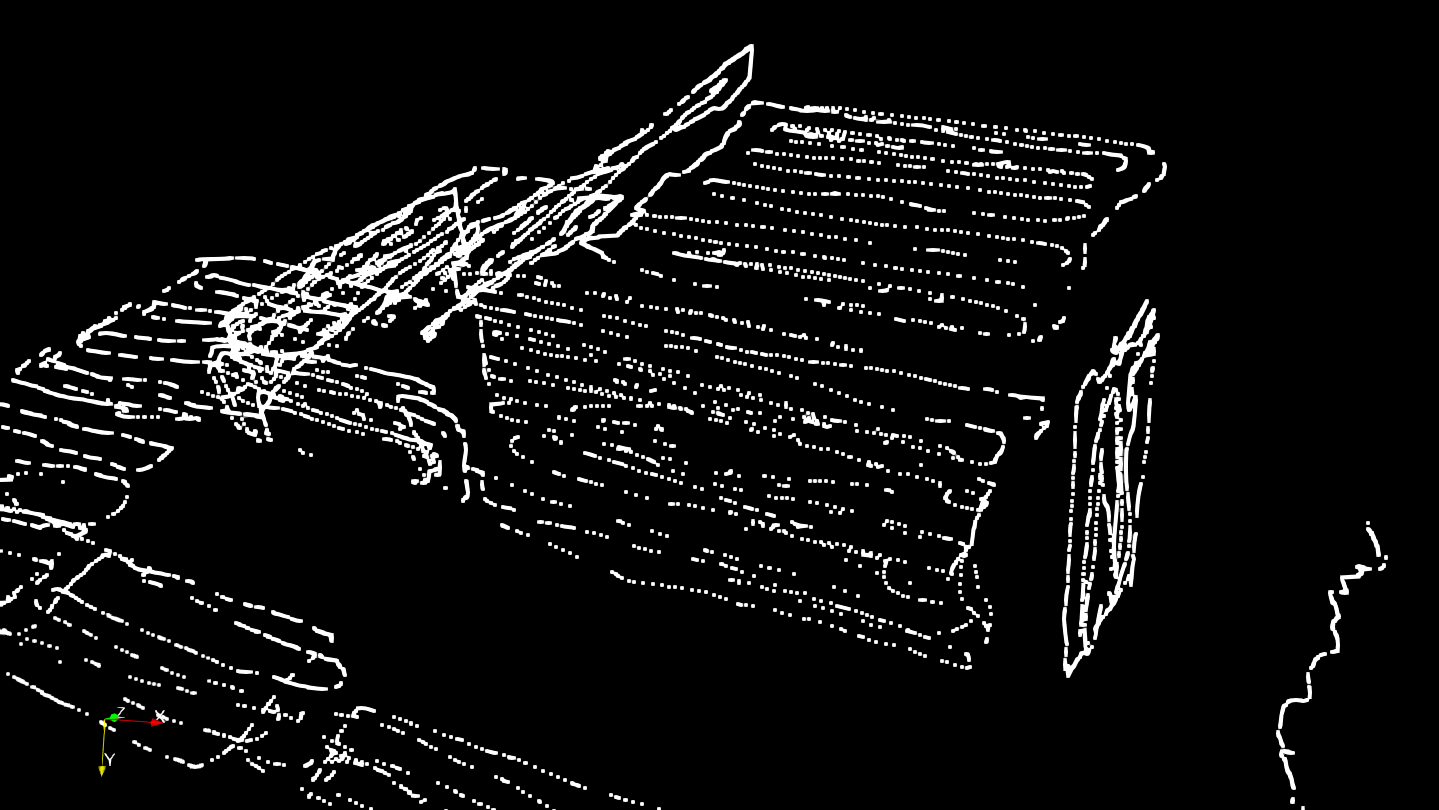}}
		\caption{Contact point cloud.}
		\label{fig:contact_cloud}
	\end{subfigure} 
	\caption{Surface shapes used for calibration (left), and the resulting dense depth (centre) and sparse contact (right) point clouds generated by KinectFusion and by contact sensing, respectively.}
	\label{fig:point_clouds}
	\vspace{-\baselineskip}
\end{figure*}

To validate our extrinsic calibration method, we performed multiple tests using the mobile manipulator shown in \cref{fig:thing}. A Kinect V2 RGB-D sensor is mounted on the sensor mast of the mobile base. The arm has a Robotiq FT300 F/T sensor attached immediately after the final wrist joint and before the gripper. For the experiments reported herein, we focus on extrinsic calibration only.

We first mapped the environment to produce a dense, fused depth map. We then used a custom impedance controller to maintain light contact between the EE and the scanned surfaces, generating a contact map. After collecting three separate RGB-D maps and three corresponding contact maps, we determined the extrinsic transform between the arm's base frame and the RGB-D sensor frame by registering each pair of point clouds. We then validated our extrinsic calibration results by performing several accuracy tests. These experiments, as well as the results, are described in greater detail below.

\subsection{Object and Surface Selection for Contact Mapping}

We demonstrated our self-calibration approach using test surfaces from two simple rectangular boxes (prisms). We chose this particular shape based on the following criteria:
	\begin{enumerate}
	\item it is representative of the types of objects readily available in many environments,
    \item its surfaces are mappable to a high fidelity by most contact or tactile sensors,
    \item the surfaces of the prism sufficiently constrain the registration solution, as discussed in Section \ref{subsec:observability}.
	\end{enumerate}
Items 1) and 2) are practical requirements based on the resolution and type of contact or tactile sensor employed. We argue that as contact and tactile sensors become more accurate and capable of producing higher-resolution measurements, these requirements may be relaxed, allowing more complex shapes to be mapped reliably.

\subsection{Depth Map Acquisition}

Depth maps of the environment were acquired with \emph{KinectFusion} \cite{Newcombe2011-tl}, taking advantage of our holonomic mobile base to generate a fused point cloud from multiple viewpoints. An example of one of the depth maps used in our experiments is shown in \cref{fig:kinect_cloud}. After collecting a full RGB-D map, we kept the mobile base fixed in its final position for the contact mapping phase. Since KinectFusion was no longer running, any base movement was not compensated for when determining the final estimate of the transform between the point clouds.

Our approach relies heavily on the accuracy of KinectFusion's mapping results (or those of any 3D mapping package). It is likely that some of the error introduced into our calibration is due to artifacts in the point cloud caused by the KinectFusion algorithm itself. Notably, KinectFusion struggles to map sharp edges and can introduce a `bow' in some planar surfaces, as shown on the right side of \cref{fig:kinect_cloud}.

\subsection{Contact Map Acquisition}

\begin{figure*}
    \centering   
    \setlength{\fboxsep}{0pt}%
    \setlength{\fboxrule}{1pt}%
    \begin{subfigure}{0.193\textwidth}
    	\fbox{\includegraphics[width=\textwidth - 2pt]{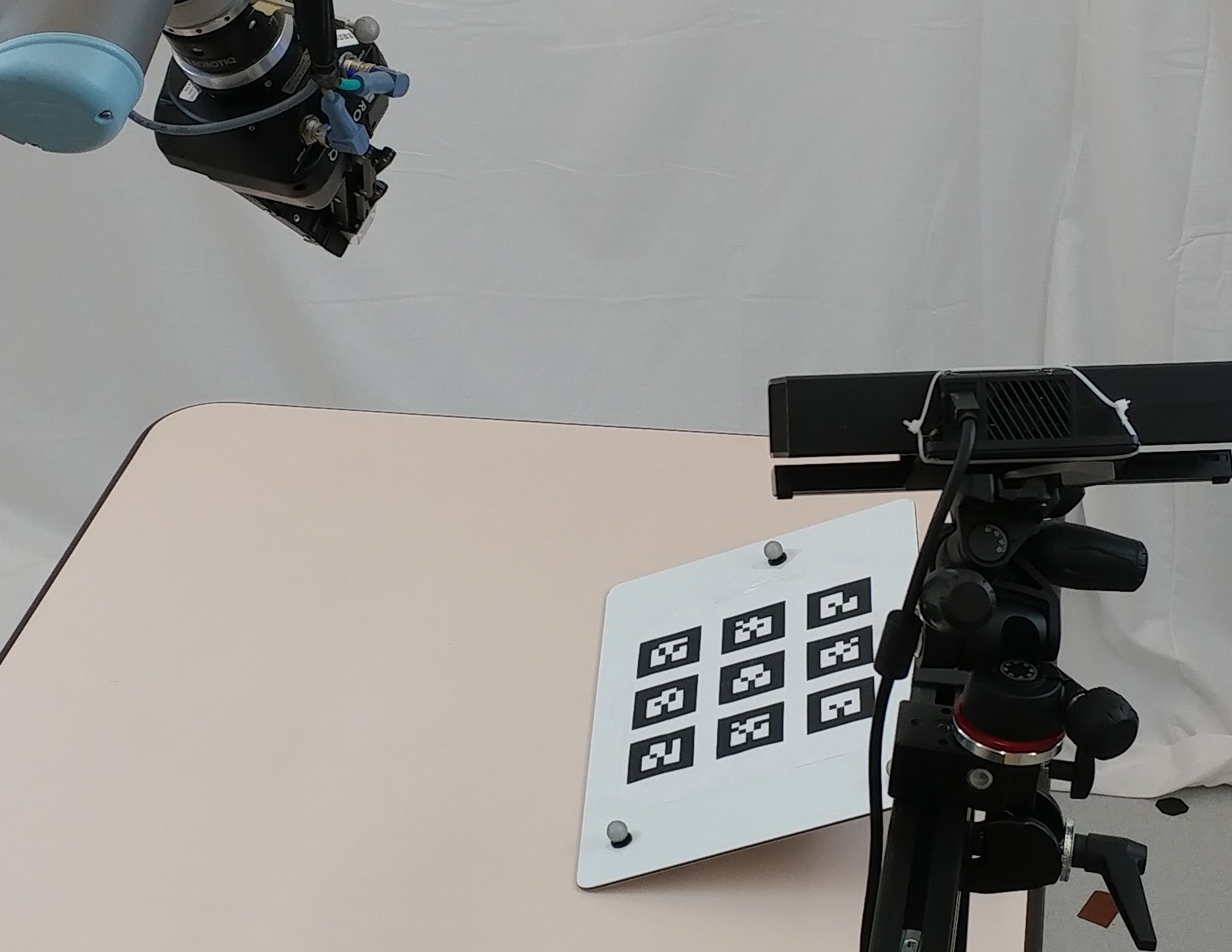}}
    \end{subfigure}
    \hfill
    \begin{subfigure}{0.193\textwidth}
       	\fbox{\includegraphics[width=\textwidth - 2pt]{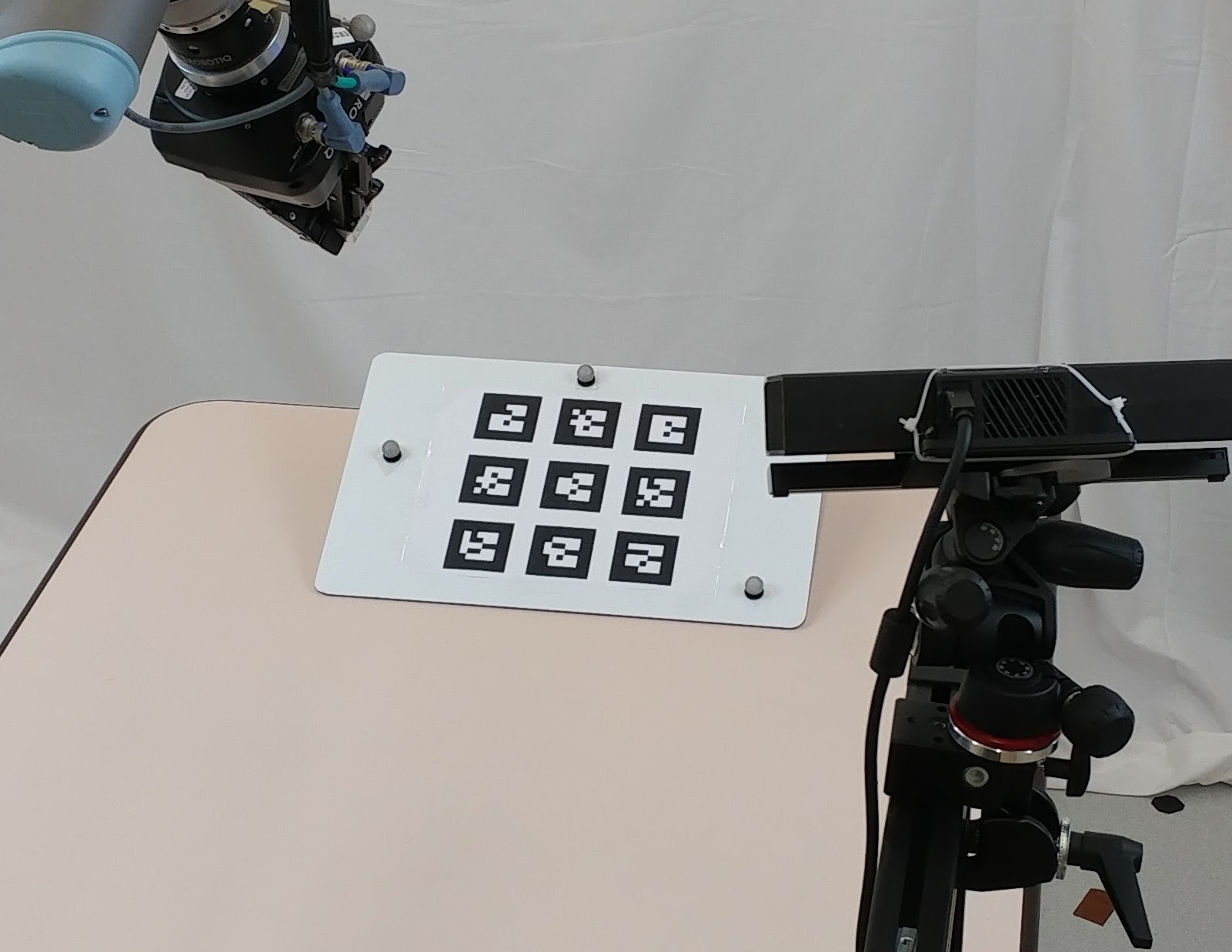}}
    \end{subfigure}
    \hfill
    \begin{subfigure}{0.193\textwidth}
    	\fbox{\includegraphics[width=\textwidth - 2pt]{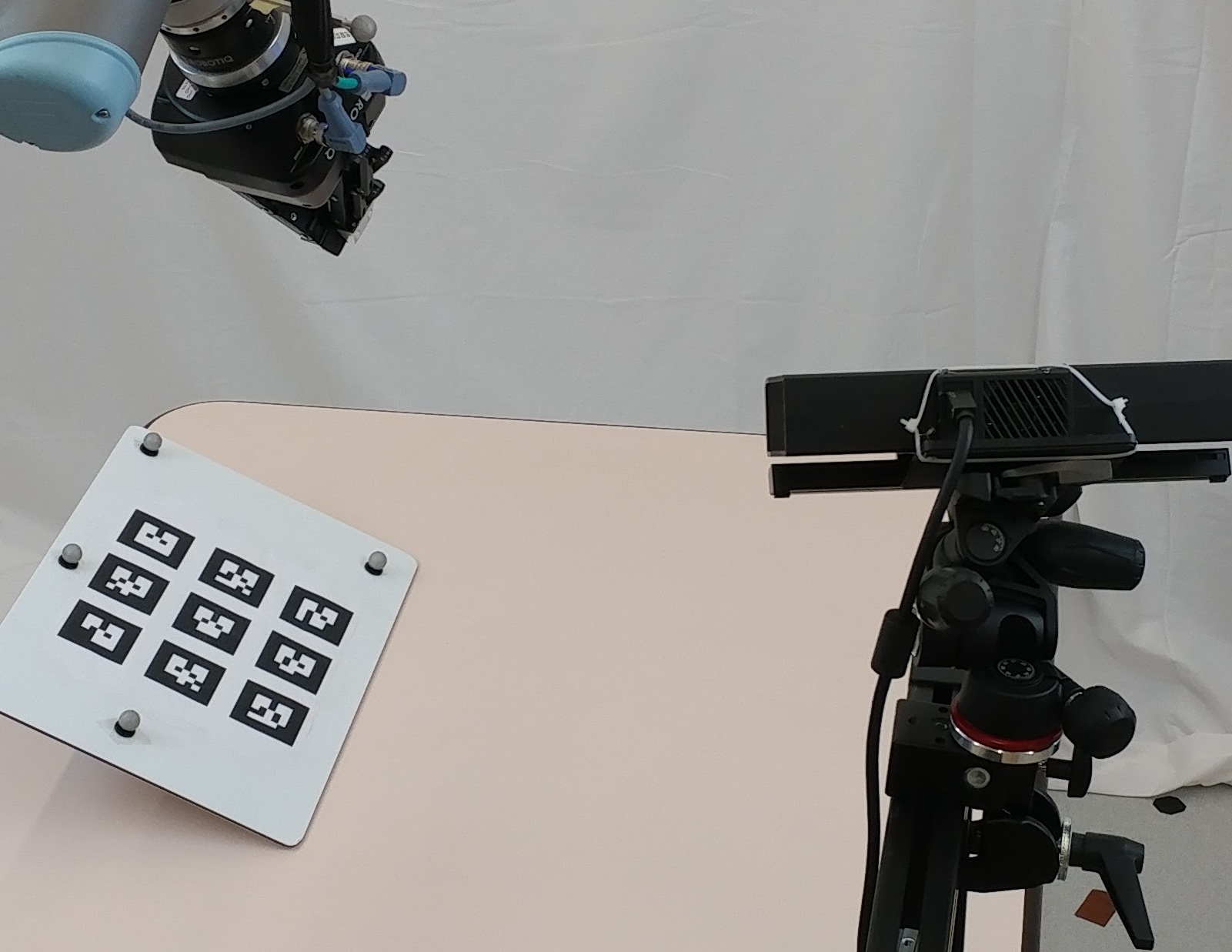}}
    \end{subfigure}
    \hfill
    \begin{subfigure}{0.193\textwidth}
       	\fbox{\includegraphics[width=\textwidth - 2pt]{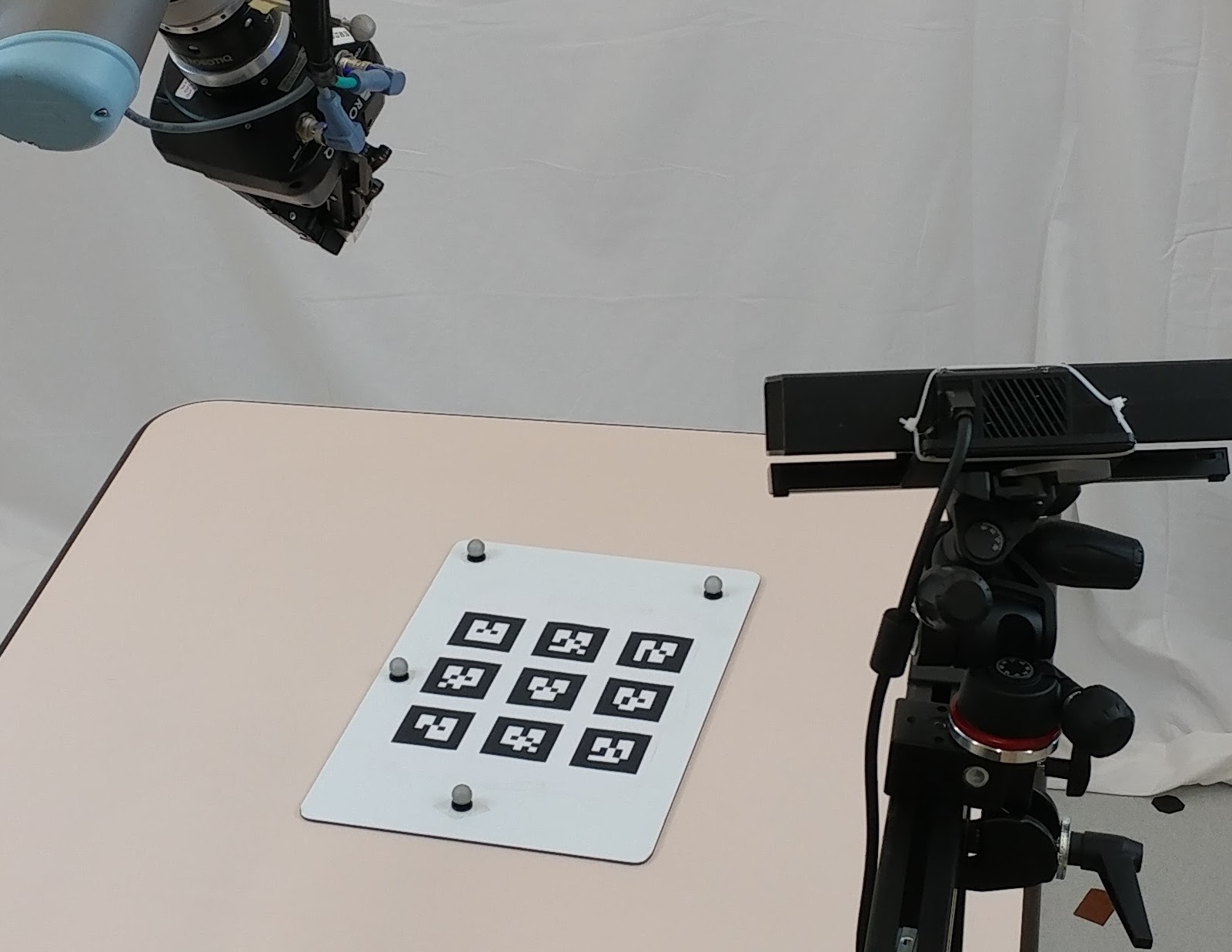}}
    \end{subfigure}
    \hfill
    \begin{subfigure}{0.193\textwidth}
       	\fbox{\includegraphics[width=\textwidth - 2pt]{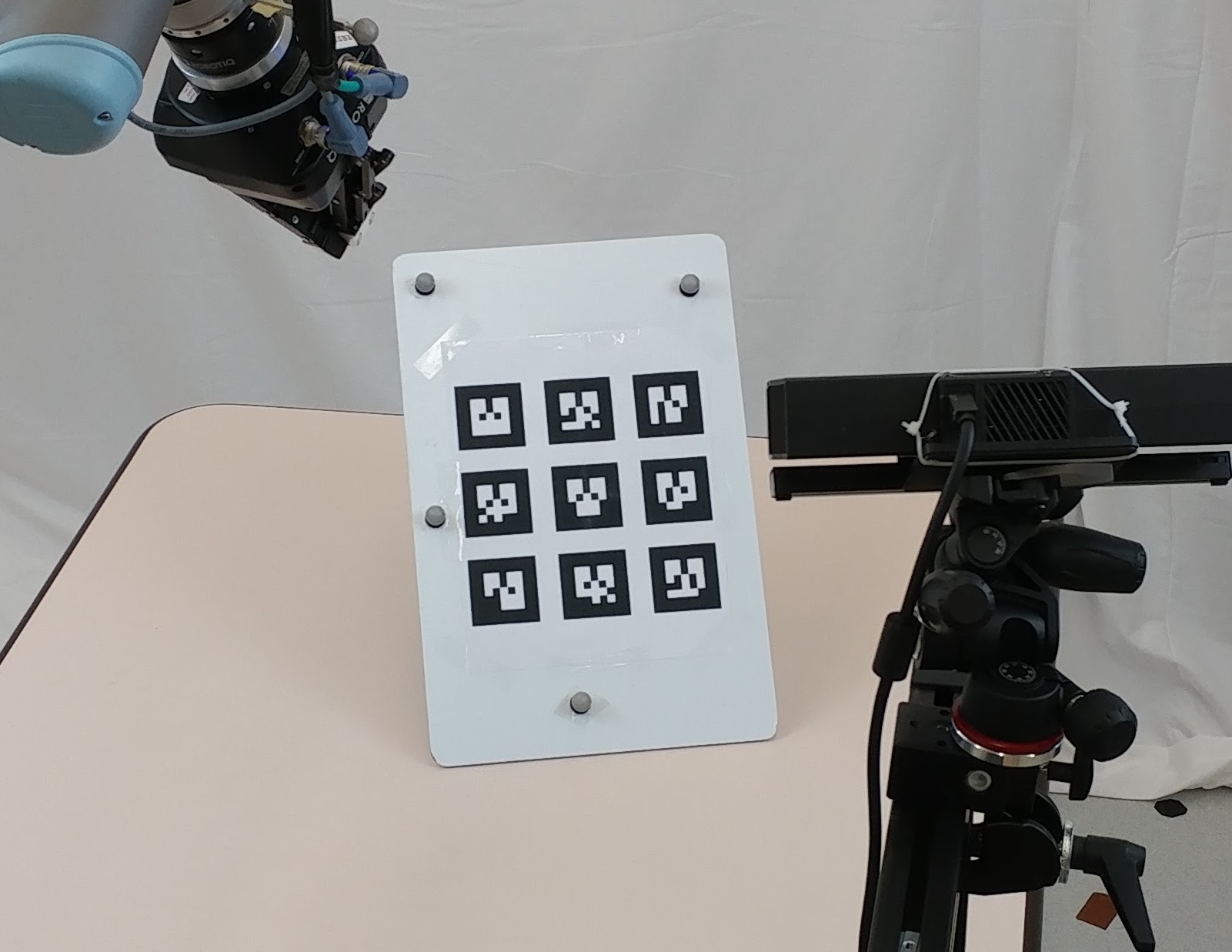}}
    \end{subfigure}     
    \caption{Five ARTag target poses used in our task-based validation procedure. We incorporated a variety of different poses to eliminate the possibility of systematic bias in our extrinsic parameter estimates.}
    \label{fig:vicon_setup}
\end{figure*}
\begin{figure}
    \centering
	\includegraphics[width=\columnwidth]{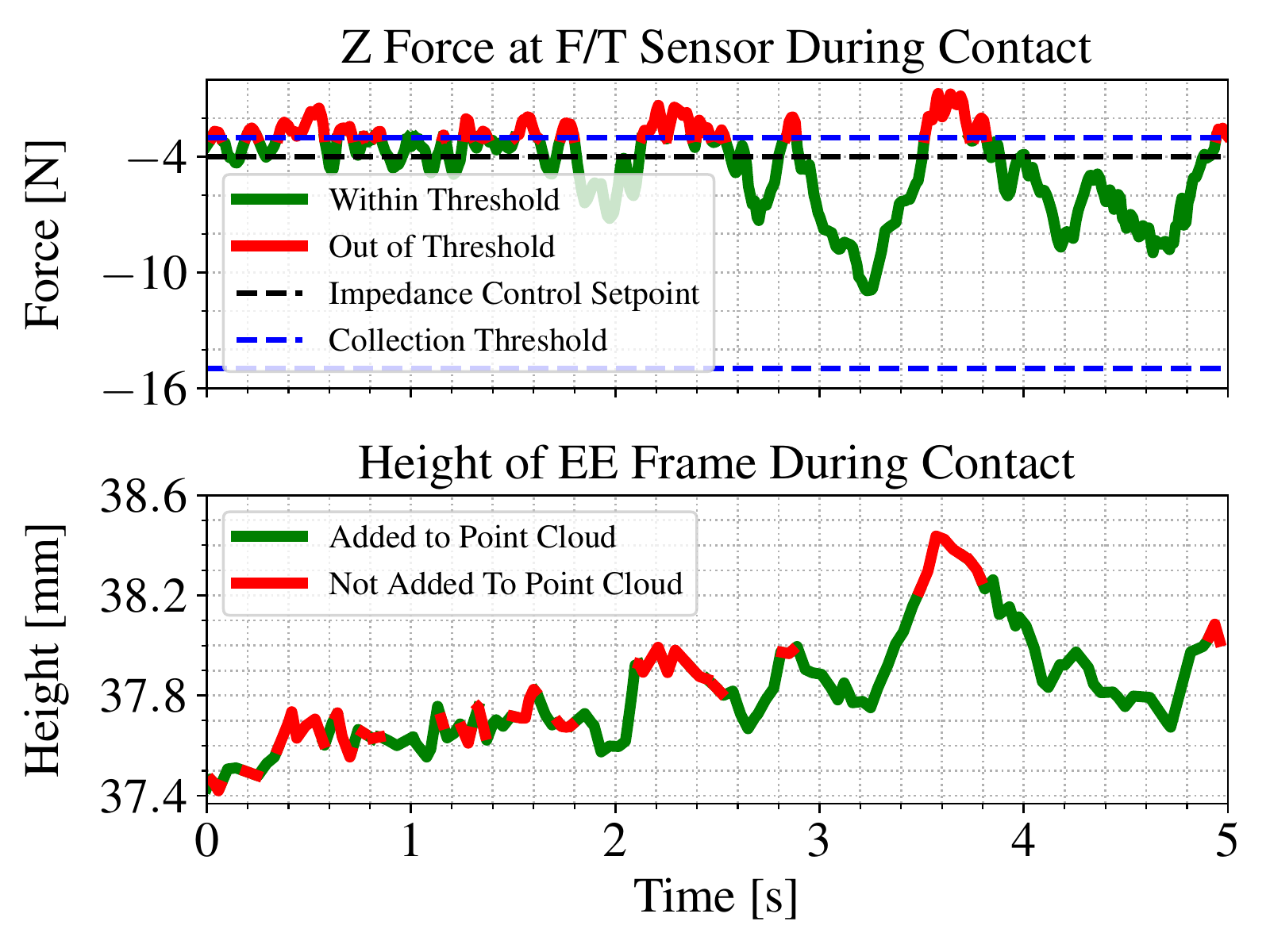}
    \caption{Force registered on the gripper, and the relative height change, between $\CoordinateFrame{E}$ and $\CoordinateFrame{B}$, measured while collecting several hundred contact points from a flat, horizontal surface. The gripper moved approximately 10 cm across the surface. Maintaining stable control given noisy force-torque measurements is difficult, as shown in the top plot. The impedance controller was able to maintain a total force on the gripper of less than 10 N. Despite the force change, the relative change in EE height never exceeded 1 mm (bottom plot).}
    \label{fig:FT_plots}
\end{figure}

To collect points for the contact map, we employed a semi-automated procedure in which the user selected the $x$ and $y$ coordinates of the EE (in the EE frame), while the $z$ position of the EE (gripper) was controlled via a PID loop to maintain light surface contact. 
In the general case, this procedure could be fully automated. We used the semi-automated procedure for convenience, and chose to leave the full motion planning problem as future work.

An example of the F/T sensor force readings and the resulting changes in height (i.e., the perpendicular distance from a scanned surface) of the EE is shown in \cref{fig:FT_plots}. The recommended threshold for contact sensing, supplied by the manufacturer of our F/T sensor, is 2 N, and the rated standard deviation of the sensor noise is 0.5 N, although we found it to be closer to 1 N in our experiments.
A typical contact map, as shown in \cref{fig:contact_cloud}, took approximately 30 minutes to collect.
The $z$-direction force reading was used as a threshold for selecting points to add to the contact point cloud. For our experiments, we set the minimum force threshold to $-3$ N (i.e., against the gripper) and the maximum force threshold to $-15$ N. The thresholds were chosen to ensure that we only collected points when there was sufficient contact and also a low risk of object deformation. The set point for the impedance controller was $-4$ N. Although intuition would suggest using an impedance value exactly in-between the threshold values, we reduced the impedance set point to ensure that the surface was not altered by contact. 

As expected from the very minimal height changes indicated in \cref{fig:FT_plots}, all of the (truly) flat, sampled surfaces do in fact appear to be flat in the contact map shown in \cref{fig:contact_cloud}. As an additional validation step, we verified the measured (34.9 cm) and known (34.5 cm) values of the distance between two parallel surfaces on one of the prisms.

\subsection{Point Cloud Registration Procedure}
\begin{table*}
  \centering
  \caption{Extrinsic calibration results from three separate trials with different manipulator trajectories.}
      \begin{tabular}{*{8}{c}}
           & & $x$ [mm] & $y$ [mm] & $z$ [mm] & $\phi$ [deg] & $\theta$ [deg] & $\psi$ [deg]
           \\\midrule
      Initial Guess && 800 & 300 & 600 & -125 & 0 & 0  \\
      Trial 1 && 839.4 & 257.3 & 676.6 & -119.07 & 1.00 & 16.23  \\
      Trial 2 && 834.6 & 259.0 & 691.6 & -120.18 & 1.27 & 15.62   \\
      Trial 3 && 836.3 & 254.0 & 695 & -120.44 & 1.38 & 14.94   \\ \midrule
      Mean (Stdev) && \textbf{836.77 (1.99)} & \textbf{256.77 (2.08)} & \textbf{687.73 (7.99)} & \textbf{-119.90 (0.59)} & \textbf{1.22 (0.16)} & \textbf{15.60 (0.53)}  \\\midrule
      \end{tabular}
  \label{tab:calib_results}
\end{table*}

\begin{figure}
	\begin{subfigure}{0.49\columnwidth}
		\includegraphics[width=\textwidth]{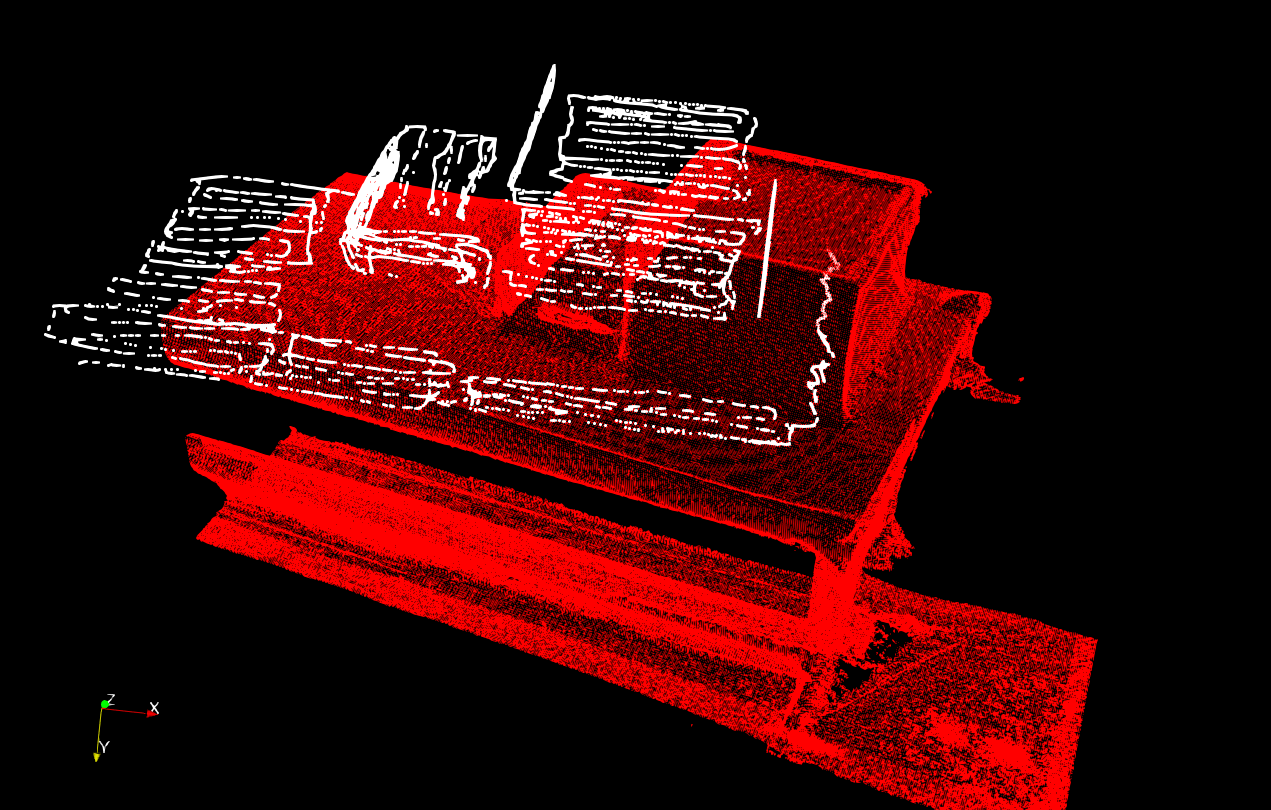}
	\end{subfigure}
	\hfill
	\begin{subfigure}{0.49\columnwidth}
		\includegraphics[width=\textwidth]{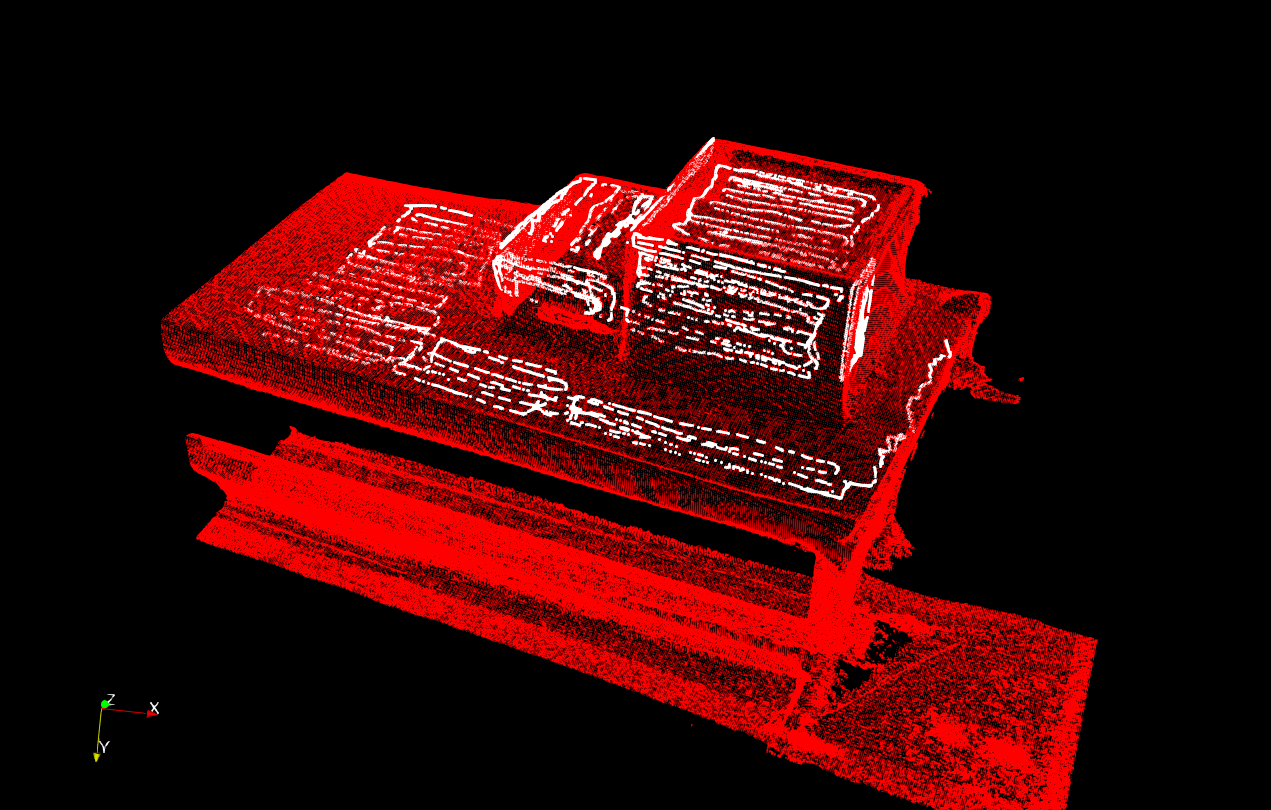}
	\end{subfigure}
	\caption{Rigid alignment of contact and depth point clouds. Left: Contact map in white with initial transform guess applied. Right: Final alignment of the point clouds after registration.}
	\label{fig:alignment}
\end{figure}

We made use of the ICP implementation available in \texttt{libpointmatcher} \cite{Pomerleau2015-rt} for rigid registration. An initial guess for the transform parameters is required; we determined this through rough hand measurement of the position and orientation of the Kinect V2 sensor relative to the manipulator base. An example of the initial and final alignments of the RGB-D and contact point clouds is shown in \cref{fig:alignment}. The final calibration results are given in \cref{tab:calib_results}. These results are the average of three separate trials. Trial I was carried out by sampling contact points from both prisms and the table, while Trials II and III were performed by sampling from the single larger prism and the table only.

\subsection{Task-Based Validation Results and Analysis}

\begin{figure}
	\centering   
	\includegraphics[width=0.99\columnwidth]{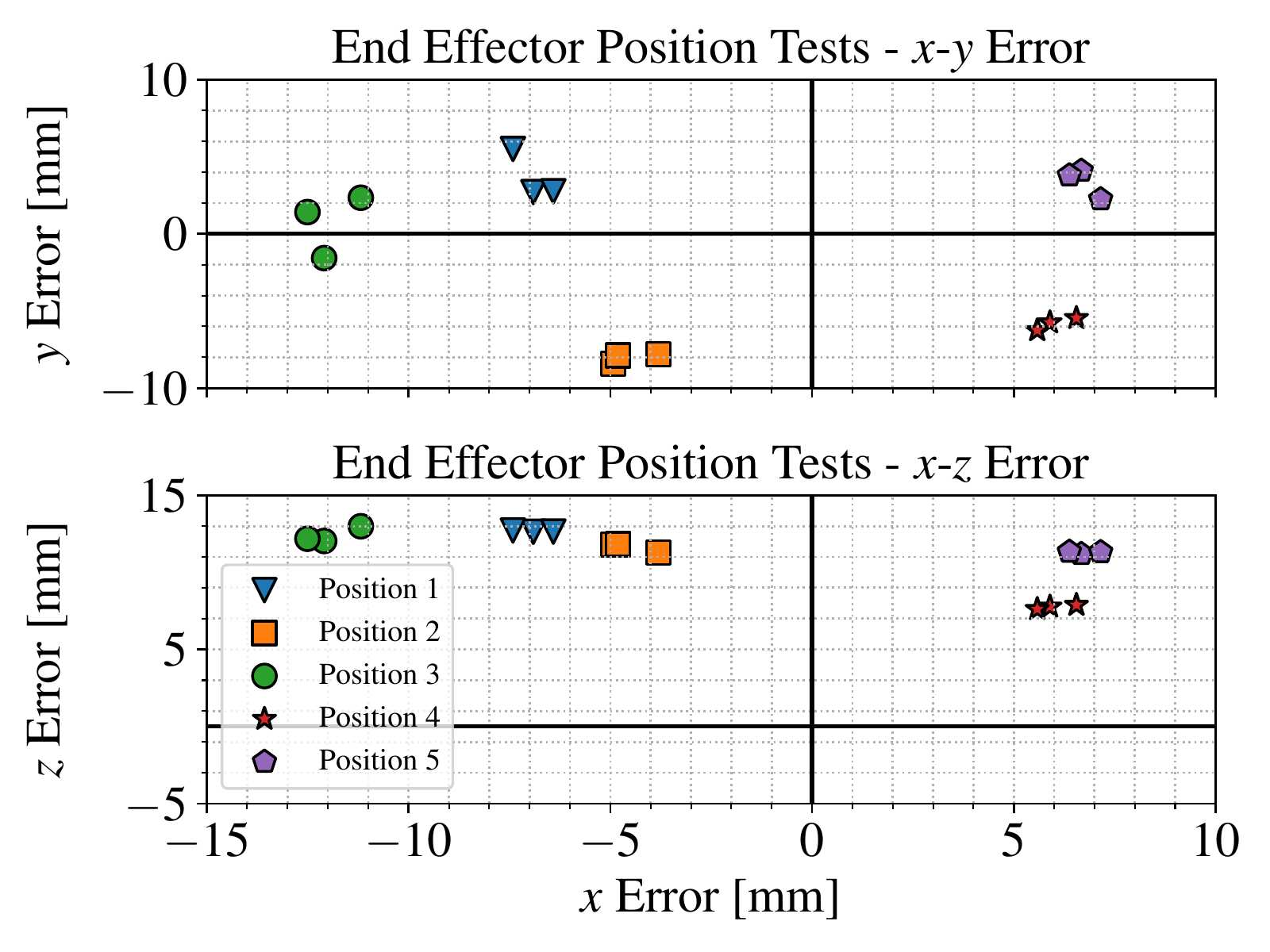}
	\caption{Errors in end effector position after the calibration procedure. We define the error to be the difference between the desired position of $\CoordinateFrame{E}$ and the actual position as measured by Vicon.}
	\label{fig:calibration_error_plots}
	\vspace{-\baselineskip}
\end{figure}

We validated our extrinsic calibration results through a task-based experiment involving a Vicon motion capture system and a planar board containing nine printed ARTags\footnote{\url{http://wiki.ros.org/ar_track_alvar}}. Vicon markers were used to track the position of the EE and the positions of the ARTags (both in the Vicon system reference frame). We placed the ARTag board in five separate poses with significant translational and rotational variation, as shown in Fig. \ref{fig:vicon_setup}. At each of these poses, we commanded the arm (specifically $\CoordinateFrame{E}$ in \cref{fig:frames}) to go to a position with a specific offset from the centre ARTag, in order to avoid a collision with the board. We recorded the translational error between the commanded positions and the ground truth positions using Vicon. The results of this experiment are visualized in \cref{fig:calibration_error_plots}. We also performed the same experiment with the initial guess for the extrinsic calibration, as shown in \cref{tab:calib_results}, but in every case, the EE missed the target position by at least 25 cm.

\begin{table}[b]
	\vspace{-0.2\baselineskip}
	\centering
	\caption{Position error for our task-based validation procedure. Each reported position is an average over three trials. The mean is calculated as the average of the absolute error.}
	\begin{tabular}{*{5}{c}}
		&& \multicolumn{3}{c}{Position Error [mm]} \\ \cline{3-5}
		&& $x$ & $y$ & $z$ \\\midrule
		Position 1 && -6.91 & 3.66 & 12.66  \\
		Position 2 && -4.51 & -8.04 & 11.64 \\
		Position 3 && -11.93 & 0.73 & 12.40 \\
		Position 4 && 6.01 & -5.82 & 7.76   \\
		Position 5 && 6.73 & 3.38 & 11.28   \\\midrule
		Mean (Absolute)&& \textbf{7.22} & \textbf{4.33} & \textbf{11.15}   \\ 
		Stdev && \textbf{7.34} & \textbf{4.82} & \textbf{1.77}   \\ \midrule
	\end{tabular}
	\label{tab:vicon}
\end{table}

The results in \cref{tab:vicon} show that the gripper position had an average total translational error of 1.4 cm. We argue that for many standard manipulator tasks, such as pick-and-place, this amount of error would not prevent a task from being completed. Additionally, it appears that there is a systematic error in the $z$ direction, which could likely be improved upon with better calibration between the KinectFusion frame and ARTag frame.

Although we did attempt to implement the non-rigid ICP solution described in Section IV with simulated joint angle biases for one of our datasets, we found that the resulting approximate Hessian had a very small determinant, making the problem unsolvable in this specific case. We suspect that this was due to the lack of variety in joint configurations commanded while collecting data, making it impossible to determine if a joint angle bias or an extrinsic calibration error was causing the final positioning error.

In Trial 1, for example, the range of the manipulator's configuration space, $\Delta\boldsymbol{\theta} = \boldsymbol{\theta}_{max}-\boldsymbol{\theta}_{min}$, covered was $\Delta\boldsymbol{\theta} = \bbm 25.6 & 3.84 & 56.4 & 170.5 & 74.5 & 17.0 \ebm^{T}$ degrees for each joint, starting from the base. The range of the task (position) space , $\Delta\mathbf{p} = \mathbf{p}_{max} - \mathbf{p}_{min}$, covered was $\Delta\mathbf{p}= \bbm 0.898 & 0.497 & 0.463 \ebm^{T}$ metres in $x$, $y$, and $z$ respectively. 
In the future, it would likely be beneficial to collect data with more variation in the arm configuration and over a larger work space, as described in \cite{Driels1990-uk}. One possibility is to add an attachment that would allow points to be sampled while relaxing the constraint that $\CoordinateFrame{E}$ be perpendicular to the contact surface; this would, however, add a separate apparatus, which we would like to avoid.

\subsection{Effects of Sampling Choice and Point Cloud Sparsification}

\begin{figure}
	\centering
	\includegraphics[width=0.99\columnwidth]{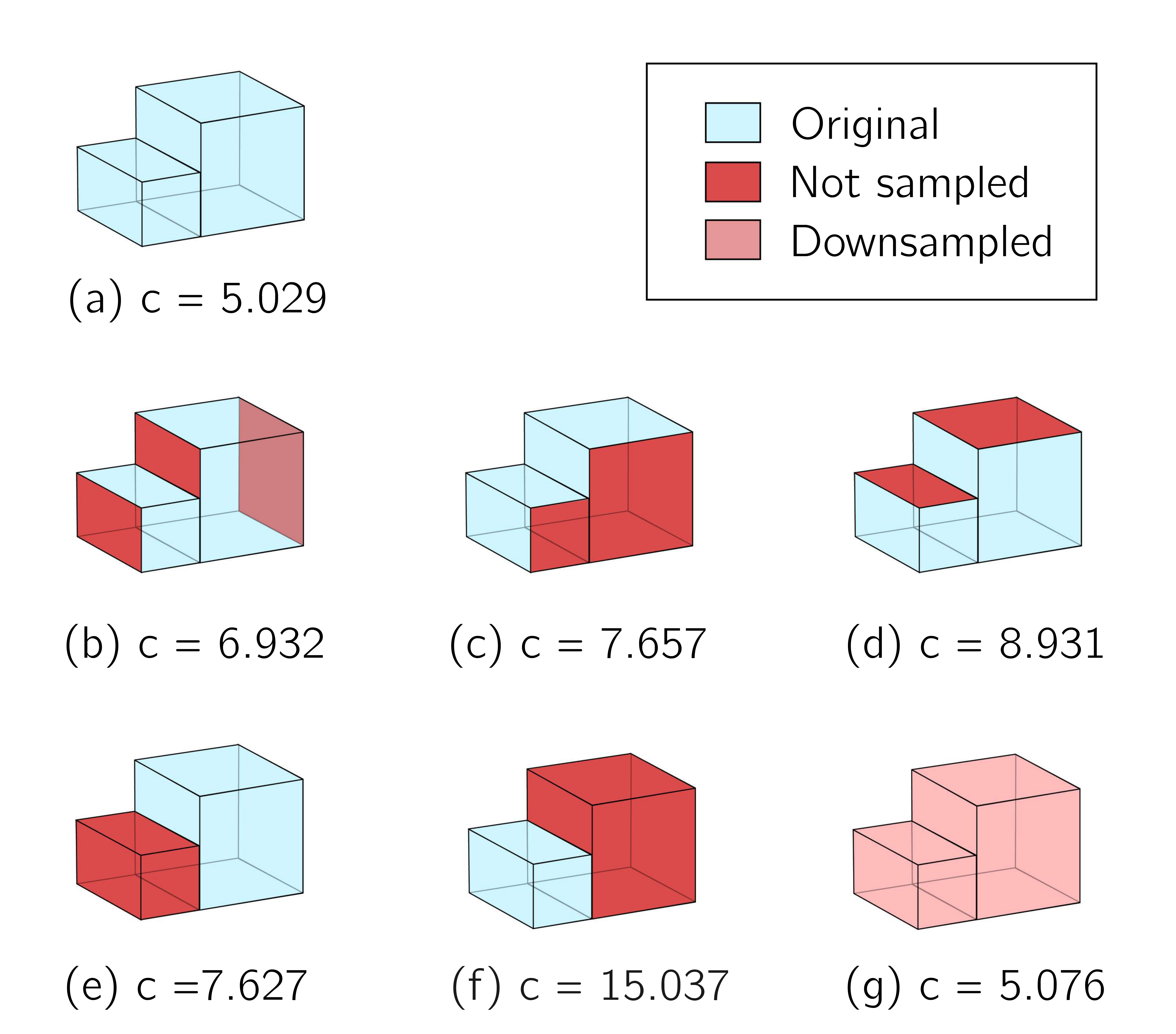}
	\caption{Stability metric $c$, defined in \cref{eq:stability_metric}, for different contact maps. Planes which were not sampled are shown in red, while the lighter red colour identifies downsampling from the original 65000 contact points to just 500 contact points.}
	\label{fig:cond_numb}
	\vspace{-\baselineskip}
\end{figure}

We studied the effects of varying the surfaces sampled to obtain the contact map and the number of contact points collected during the calibration procedure. From a practical point of view, calibrating using simple shapes and fewer contact points potentially increases both the flexibility and speed of the process. We computed the stability metric, given by \cref{eq:stability_metric}, of the converged solution under different sampling scenarios, as shown in \cref{fig:cond_numb}. The choice of sampled surfaces is related to the stability of the converged solution, as expected.

Our original contact map converged with a stability measure of $c=5.02976$, where a larger $c$ value implies a \emph{less} stable convergence. We first studied the effect of \emph{not} sampling from certain planes, as shown in Fig. \ref{fig:cond_numb}\textcolor{red}{b}, \ref{fig:cond_numb}\textcolor{red}{c} and \ref{fig:cond_numb}\textcolor{red}{d}, which increased the $c$ values to 6.932, 7.657, and 8.931, respectively. Additionally, not sampling any points on the left or right prisms, as shown in Fig. \ref{fig:cond_numb}\textcolor{red}{e} and \ref{fig:cond_numb}\textcolor{red}{f}, increased the $c$ values to 7.628 and 15.037, respectively.
\begin{figure}
	\centering
	\includegraphics[width=0.99\columnwidth]{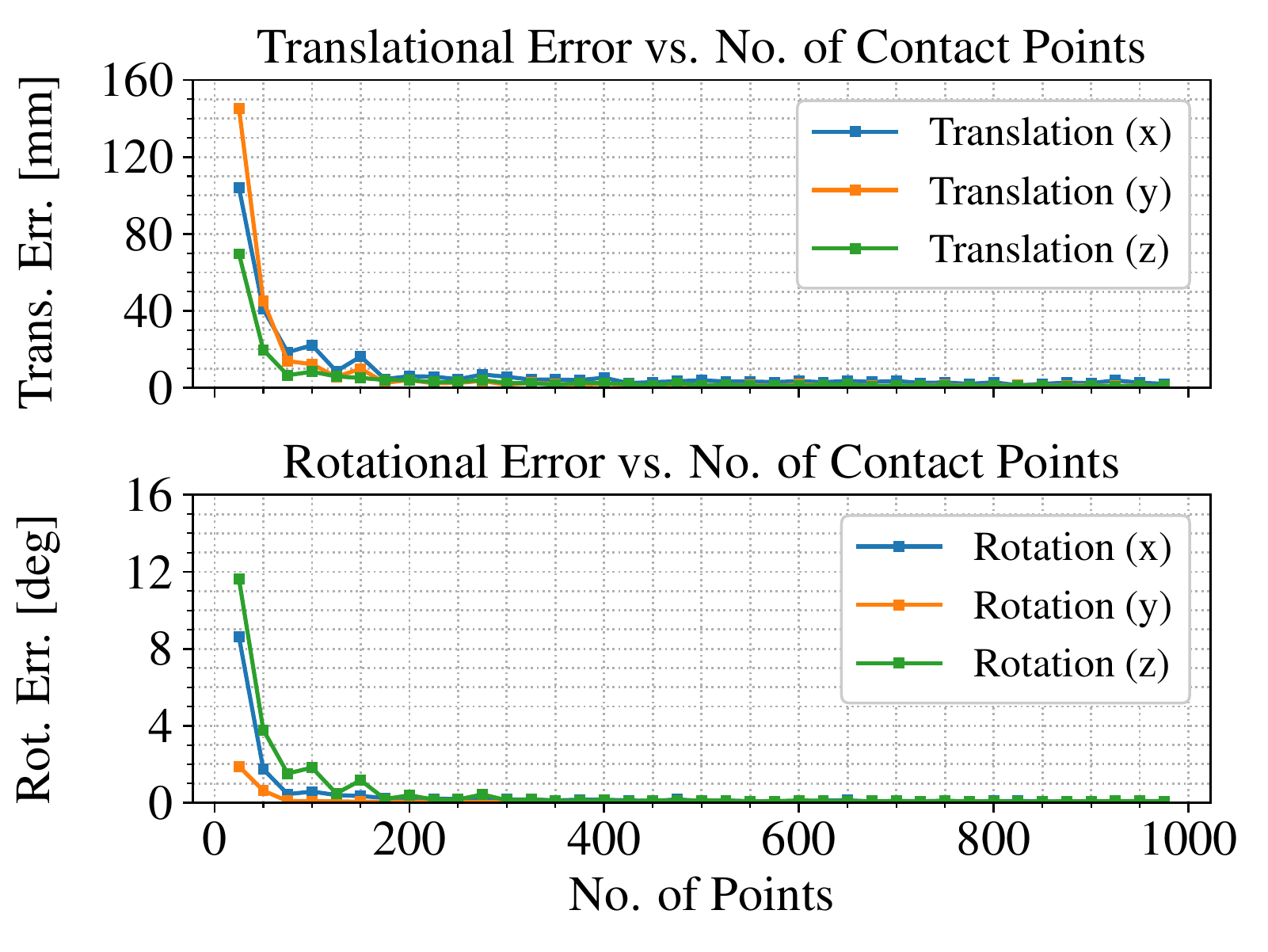}
	\caption{Extrinsic transform error as a function of the number of contact points sampled. Each point in the graph represents the average error for 20 randomly sampled subsets of contact points. Below a certain threshold, the solution becomes unstable and diverges.}
	\label{fig:downsampling_plots}
	\vspace{-\baselineskip}
\end{figure}

On the other hand, downsampling from 65,000 to 500 contact points, as visualized in \cref{fig:cond_numb}\textcolor{red}{g}, had a limited effect on the stability of the solution. Specifically, we uniformly downsampled by randomly selecting a certain number of points from the original contact map. This shows that there were many redundant points and that a sparser cloud could likely have been used. \cref{fig:downsampling_plots} demonstrates the effect of  downsampling even further; as long as a minimum density and distribution of points is maintained in the contact map, the procedure converges reliably and stably. The key factor to consider to obtain a stable solution is the variety of surfaces sampled. Note that these results hold for rigid registration---however, in the non-rigid case, increasing the number of sampled points is likely to improve the quality of the solution.

\section{Conclusions}

We have presented a novel, proof-of-concept method to self-calibrate the extrinsic transform between a manipulator EE and a fixed depth camera, by leveraging contact sensing as a previously unused modality for this application. We validated our calibration results through a task-based test performed in a Vicon motion capture facility, demonstrating $\sim$1 cm positioning accuracy post-calibration. The proposed method only uses sensors that are readily available for most standard manipulators and does not rely on any fiducial markers or bulky and costly external measurement devices.

Future work includes evaluating the use of sparse point cloud registration \cite{Srivatsan_undated-lp} to reduce the quantity of contact points needed to ensure convergence, implementing the full non-rigid calibration procedure to recover the DH  parameters (using a data set with a wider variety of manipulator configurations), as well as examining the use of higher-fidelity contact or tactile sensors. We are also exploring methods to produce more accurate depth maps to further improve calibration accuracy. Finally, we are working to extend the method to incorporate extrinsic calibration of monocular cameras, where the unknown scale parameter can potentially be folded in as a calibration parameter. 
\balance
\bibliographystyle{IEEEcaps}
\bibliography{refs-abbrev}
\end{document}